\documentclass[twoside,11pt]{article}

\usepackage{blindtext}

\usepackage[utf8]{inputenc} %
\usepackage[T1]{fontenc}    %
\usepackage{url}            %
\usepackage{booktabs}       %
\usepackage{amsfonts}       %
\usepackage{nicefrac}       %
\usepackage{microtype}      %
\usepackage[dvipsnames]{xcolor}         %
\usepackage{bm}             %
\usepackage{wrapfig}        %
\usepackage{varwidth}      %
\usepackage{amsmath, amssymb, amsfonts}
\usepackage{graphicx}

\usepackage{xspace}
\usepackage{algorithm}
\usepackage{algpseudocode}

\usepackage{multirow}

\usepackage{subcaption}
\usepackage{placeins}

\DeclareMathOperator{\argmax}{argmax}
\newcommand{\R}{\mathbb{R}}
\newcommand{\normal}{\mathcal{N}}

\newcommand{\fedavg}{\textsc{FedAvg}\xspace}
\newcommand{\fedsgd}{\textsc{FedSGD}\xspace}
\newcommand{\fedavgm}{\textsc{FedAvgM}\xspace}
\newcommand{\fedprox}{\textsc{FedProx}\xspace}
\newcommand{\fedopt}{\textsc{FedOpt}\xspace}
\newcommand{\fedoptround}{\textsc{FedOptRound}\xspace}
\newcommand{\fad}{federated AD\xspace}
\newcommand{\Fad}{Federated AD\xspace}

\newcommand{\perfedavg}{\textsc{PerFedAvg}\xspace}

\newcommand{\serverupdate}{\textsc{ServerUpdate}\xspace}

\newcommand{\clientupdate}{\textsc{ClientUpdate}\xspace}

\newcommand{\SERVER}{\texttt{@SERVER}\xspace}
\newcommand{\CLIENTS}{\texttt{@CLIENTS}\xspace}
\newcommand{\fedbroadcast}{\texttt{federated\textunderscore broadcast}\xspace}
\newcommand{\fedsum}{\texttt{federated\textunderscore sum}\xspace}
\newcommand{\fedmean}{\texttt{federated\textunderscore mean}\xspace}
\newcommand{\fedselect}{\texttt{federated\textunderscore select}\xspace}

\newif\ifdraft
\drafttrue

\definecolor{darkgreen}{rgb}{0,0.4,0.0}
\definecolor{darkblue}{rgb}{0,0.0,0.4}
\definecolor{bcolor}{rgb}{0.6,0.0,0.6}
\ifdraft
\newcommand{\zctodo}[1]{[{\color{red}{\textbf{ZC TODO:} \emph{#1}}}]}
\newcommand{\zgtodo}[1]{[{\color{darkgreen}{\textbf{ZG TODO:} \emph{#1}}}]}
\newcommand{\ktodo}[1]{[{\color{darkblue}{\textbf{K TODO:} \emph{#1}}}]}
\newcommand{\btodo}[1]{[{\color{bcolor}{\textbf{BM:} \emph{#1}}}]}
\else
\newcommand{\zctodo}[1]{}
\newcommand{\zgtodo}[1]{}
\newcommand{\ktodo}[1]{}
\newcommand{\btodo}[1]{}
\fi

\usepackage{tabularx}
\usepackage{adjustbox}
\usepackage{tikz}
\usetikzlibrary{decorations.pathmorphing}
\usetikzlibrary{arrows}
\usetikzlibrary{positioning}

\newcommand{\largemodel}{\ensuremath{n}}
\newcommand{\smallmodel}{\ensuremath{m}}

\usepackage{jmlr2e}

\usepackage[capitalise]{cleveref}

\newtheorem{class}{Class}
\newtheorem{question}{Question}

\usepackage{lastpage}
\jmlrheading{25}{2024}{1-\pageref{LastPage}}{2/23; Revised 4/24}{11/24}{23-0223}{Keith Rush and Zachary Charles and Zachary Garrett}

\ShortHeadings{Federated Automatic Differentiation}{Rush, Charles, and Garrett}

\firstpageno{1}

\begin{document}

\title{Federated Automatic Differentiation}

\author{\name Keith Rush \email krush@google.com \\
       \addr Google Research \\
       Seattle, WA, USA
       \AND
       \name Zachary Charles \email zachcharles@google.com \\
       \addr Google Research\\
       Seattle, WA, USA
       \AND
       \name Zachary Garrett \email zachgarrett@google.com \\
       \addr Google Research\\
       Seattle, WA, USA}

\editor{Dan Alistarh}

\maketitle

\begin{abstract}%
Federated learning (FL) is a framework for learning across an axis of group partitioned data (heterogeneous clients) while preserving data privacy, under the orchestration of a central server.
FL methods often compute gradients of loss functions purely locally (e.g. at each client), typically using automatic differentiation (AD) techniques.
In this work, we consider the problem of applying AD to federated computations while preserving compatibility with privacy-enhancing technologies.
We propose a framework, federated automatic differentiation (\fad), that 1) enables computing derivatives of functions involving client and server computation as well as communication between them and 2) operates in a manner compatible with existing federated technology.
We show, in analogy with AD, that \fad may be implemented using various accumulation modes, which introduce distinct computation-communication trade-offs and systems requirements.
Further, we show that a broad class of federated computations is \emph{closed} under these modes of \fad, implying that if the original computation can be implemented using privacy-preserving primitives, its derivative may be computed using the same primitives.
We then show how \fad can be used to create algorithms that dynamically learn components of the algorithm itself. We demonstrate that performance of \fedavg-style algorithms can be significantly improved by using \fad in this manner.
\end{abstract}


\section{Introduction}\label{sec:intro}

Federated learning (FL) is a paradigm for distributed learning in which clients collaboratively learn without directly sharing data. When combined with formal privacy mechanisms, it enables learning across decentralized data sources while preserving the privacy of the clients. FL encompasses a variety of settings with widely varying system characteristics. Two particularly noteworthy settings are \emph{cross-device FL}, characterized by its many lightweight, unreliable clients, and \emph{cross-silo FL}, characterized by a moderate number of reliable clients. We defer to \citet{aop} for a more detailed introduction and taxonomy.

FL has seen notable success in production systems and applications~\citep{NWP18, gboard_18, fl_at_scale, papaya, apple_fl, keyword_speech, oov_fl, emoji_fl}.
Despite this progress, developing and deploying effective FL methods remains difficult. When data is heterogeneous, FL methods can fail to converge to critical points of the underlying loss, or may not converge at all~\citep{local_fixed_points_kaust, fedsplit}. FL methods often use multiple local training steps per communication round to reduce total communication~\citep{mcmahan2017communication}, which can dictate a trade-off between initial convergence and accuracy~\citep{charles_and_konecny}. This algorithmic pattern means that many FL methods are not equivalent to (stochastic) gradient descent on any loss function~\citep{charles_and_rush}. These and related observations have inspired a slew of work that attempts to improve convergence via techniques such as learning rate decay~\citep{fedchain} and control variates used to mitigate ``client drift''~\citep{scaffold, fedlin, proxskip}.

Unfortunately, such methods often require significant theoretical insight to design, may not work in cross-device FL settings~\citep{proxskip}, fail to perform well in practice~\citep{fedlin}, or may be incompatible with formal privacy techniques such as differential privacy and secure aggregation, which can limit what algorithms are possible to execute (see \citealt[Chapter 4]{aop}). They can also be difficult to tune. Hyperparameter tuning in FL can be prohibitively complex due to things like data access restrictions, system constraints, and the presence of many hyperparameters~\citep{fed_hparam_tuning}. Moreover, as discussed above, FL methods that use fixed hyperparameters may face convergence issues regardless of tuning~\citep{charles_and_konecny}.

In this work, we consider the problem of how to make it easier to design, implement, and dynamically adjust FL algorithms and their hyperparameters during training. We are motivated by the success of automatic differentiation (AD) in enabling rapid development of new algorithms and techniques in machine learning (see \citep{autodiff_survey} for a thorough survey). The development of ML-focused AD frameworks, such as JAX~\citep{jax}, TensorFlow~\citep{tensorflow}, and PyTorch~\citep{pytorch}, has accelerated this, enabling easy differentiation of loss functions with respect to user-specified parameters. We wish to enable such ease-of-use functionality and rapid experimentation in FL settings.

We are particularly interested in using AD to dynamically adjust FL algorithms in tandem with training. While such methods can fail due to perturbation confusion~\citep{siskind2005perturbation, manzyuk2019perturbation} and convergence issues~\citep{christianson1994reverse, habiba2021neural}, in practice they can often yield promising results when applied to neural network training. For example, AD has been applied to learned optimizers and update rules~\citep{learning_synaptic_update,Schmidhuber1994OnLH,learning2learn,wichrowska17a,li17a,li17b,lv17,bello17,metz19a,blur, vicol21a, velo_metz}, neural architecture search~\citep{elsken2019neural, zoph2016neural}, meta-learning~\citep{maml, blur, meta_learning_unsupervised, meta_embed, pham2018efficient}, and learned compressors~\citep{Oktay2020Scalable}. One noteworthy example of this dynamic adjustment of algorithms is \emph{hypergradient descent} which applies (stochastic) gradient descent to optimizer hyperparameters, such as the learning rate, in tandem with training~\citep{bengio_grad_hparam_opt, hypergrad_descent}.

Hypergradient descent has seen recent investigation in FL settings, due in part to the aforementioned difficulties of hyperparameter tuning. For example, the exponentiated gradient descent mechanism proposed by \citep{fed_hparam_tuning} is a form of numerical differentiation with respect to hyperparameters, \citet{fed_hypergrad_descent} apply approximate hypergradient descent to certain hyperparameters, and \citet{fednest} use hypergradient approximations for a class of bi-level federated optimization problems.~\citet{wang2023fedhyper} derive specific hypergradient descent algorithms for FL methods\footnote{We note that this appeared after the first version of this work.}. While these methods are theoretically-justified and empirically effective, they rely on manually-derived gradient formulas and approximations. The process of computing such quantities is often time-consuming, error-prone, and potentially inefficient from an implementation perspective~\citep{autodiff_survey}. 

\subsection{Contributions}
In this work, we seek to unify FL research with advanced AD by developing a framework for \emph{federated automatic differentiation}, or \fad. This framework enables the computation of \emph{exact} gradients through general federated computations,\footnote{In short, these are computations on data that have an explicit notion of where the data resides in a federated system. We will discuss these in detail in \cref{sec:fed_comp}.}
potentially involving multiple communication rounds between clients and server.
Notably, this framework allows us to differentiate arbitrary outputs with respect to arbitrary (server-side) inputs, without relying on hand-computed derivative formulas, numerical derivatives, or gradient approximations. Like AD, it can be applied to computations involving programming constructs such as recursion and branching.\footnote{Since the initial version of this work appeared, we have developed an open-source library that implements \fad in JAX. See \citep{rush2024fax} for details.}

Federated AD operates by augmenting a federated computation with a component that computes its derivative. Just as AD has different implementations (or ``modes''), we give three primary modes for \fad (forward-, reverse-, and mixed-mode). While the three modes have different system and communication overheads and requirements (which we discuss in detail), they enable exact gradient computation with only a constant factor of computation overhead. Moreover, we show that for a large class of computations, each of these modes preserves compatibility with formal privacy mechanisms, including differential privacy and secure aggregation.

After building up the \fad framework, we show how it can be used to perform federated hypergradient descent in tandem with the popular \fedopt method~\citep{afo}. In contrast to prior work~\citep{fed_hparam_tuning, fed_hypergrad_descent, fednest} \fad enables applying hypergradient descent to both client hyperparameters and server hyperparameters. We then specialize this methodology to 1) learning server optimizer hyperparameters and 2) learning client weighting schemes. We apply the resulting methods to a variety of benchmark FL tasks. We find that the resulting method can exhibit significantly better convergence properties than ``static'' \fedopt, and its ability to dynamically adjust the hyperparameters over time plays a crucial role in this behavior.

\subsection{Goals and Limitations} 
Our goal is to begin the development of a framework and set of tools that makes it easier to develop dynamic algorithms for FL. While we apply \fad to derive algorithms for federated hypergradient descent, we stress that this is a relatively simple use of \fad. We focus on 1) showcasing the utility of \fad in enabling dynamically learned parameters and 2) showing how \fad enables powerful new classes of FL methods.
We do not attempt to demonstrate that this application of \fad outperforms related methods (such as the methods of \citep{wang2023fedhyper}).
Rather, we attempt to motivate, sketch, and enable research in increased dynamism for FL algorithms by the introduction of a new tool in the FL researcher's repertoire.
This framework also enables a substantial simplification of many prior methods, for example by reducing the need for gradient approximations.
By way of analogy, backpropagation has served as a fundamental enabler to the dramatic development of centralized ML by enabling efficient and flexible derivative computations, while also simplifying model and algorithm expression.
We seek to mirror this in the younger field of FL.

\section{Preliminaries and Background}

Throughout this work, we assume that all relevant functions are differentiable. The results all hold with sub-differentiable functions as well, but we restrict to differentiability for notational simplicity. In the remainder of this section, we provide sufficient background on various topics in order to give a fully-featured but succinct description of \fad. We do not attempt to give a complete overview of topics, and give references as needed.

\subsection{Automatic Differentiation}\label{sec:auto_diff}

Automatic differentiation (AD), also known as ``autodiff'', is a family of techniques used to efficiently compute derivatives of numeric functions expressed as computer programs. While we defer the interested reader to works such as \citep{autodiff_survey} for a more complete picture, we will discuss some background on AD that will serve as grounding for the remainder of this work.

AD methods typically augment function evaluation with auxiliary derivative computations. This is done by reducing programs to a set of ``elementary'' operators whose derivatives are known, and applying the chain rule to accumulate and combine these derivatives. Because this is done by decomposing the program itself, it can be applied to functions involving complex programming constructs, leveraging the observation that ``AD is blind with respect to any operation, including control flow statements, which do not alter numeric values''~\citep[Section 3]{autodiff_survey}.

\begin{figure}[t]
\caption{Computational graph of $f(x_1, x_2) = \sin(x_1) + x_1x_2$. Here, $v_i$ is an intermediate value of the computation as given in \eqref{eq:basic_function}.}
\includegraphics[width=0.6\linewidth]{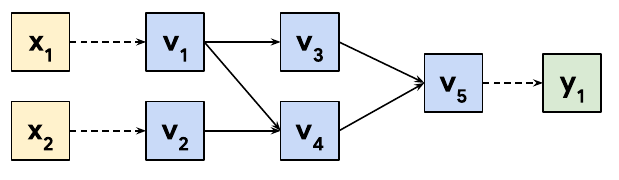}
\centering
\label{fig:autodiff_example}
\end{figure}

We can represent functions as \textbf{computational graphs}~\citep{computational_graph}. Using the notation of \citet{principles_of_autodiff}, we represent the input variables as $x_1, \dots, x_n$ and the outputs as $y_1, \dots, y_m$. The inputs and outputs are linked by a directed acyclic graph (DAG) with intermediate nodes $v_1, \dots, v_l$, where each intermediate node is some elementary function of its parents in the graph.

For example, $y_1 = \sin(x_1) + x_1x_2$ can be computed via elementary operations as follows:
\begin{equation}\label{eq:basic_function}
\begin{split}
    v_1 &= x_1\\
    v_2 &= x_2\\
    v_3 &= \sin(v_1)\\
    v_4 &= v_1v_2\\
    v_5 &= v_3 + v_4\\
    y_1 &= v_5
\end{split}
\end{equation}
This has a computational graph given in \cref{fig:autodiff_example}. Note that the dashed lines here indicate equality (that is, $v_1 = x_1, v_2 = x_2, v_5 = y_1$). Suppose we wish to compute the derivative of $y_1$ with respect to $x_1$. Two standard AD techniques for this are \textbf{forward-mode} and \textbf{reverse-mode} auto-differentiation. In forward-mode, we move forward in the graph starting from $x_1$. As we compute each intermediate node $v_i$, we also compute
\begin{equation}\label{eq:forward_mode}
\Dot{v}_i = \dfrac{\partial v_i}{\partial x_1}
\end{equation}
via the chain rule, using the fact that we know $v_j$, $\Dot{v}_j$ for $j < i$. For example, applying the chain rule to \eqref{eq:basic_function},
\begin{align*}
    \dot{v}_5 = \dfrac{\partial(v_3 + v_4)}{\partial x_1} = \dot{v}_3 + \dot{v}_4.
\end{align*}

In reverse-mode, we first compute a ``forward pass'' through the graph to compute all intermediate $v_i$. We then move backwards through the graph, starting at $y_1$. At each node we compute
\begin{equation}\label{eq:reverse_mode}
\overline{v}_i = \dfrac{\partial y_1}{\partial v_i}
\end{equation}
using the chain rule, using the fact that we know $v_j$, $\overline{v}_j$ for $j > i$. For example, applying the chain rule to \eqref{eq:basic_function},
\begin{align*}
    \overline{v}_1 &= \dfrac{\partial y_1}{\partial v_1}\\
    &= \dfrac{\partial y_1}{\partial v_3}\dfrac{\partial v_3}{\partial v_1} + \dfrac{\partial y_1}{\partial v_4}\dfrac{\partial v_4}{\partial v_1}\\
    &= \overline{v}_3\cos(v_1) + \overline{v}_4\dfrac{v_4}{v_1}.
\end{align*}

Both methods generalize to compute the Jacobian of a vector-valued function. The difference between them corresponds to choosing an order for multiplications of Jacobian matrices specified by the chain rule, with `forward-mode' corresponding to left-to-right multiplication and `backwards mode' right-to-left. If $f: \R^n \to \R^m$, then forward-mode requires $n$ forward passes to compute this Jacobian (one for each input), while reverse-mode requires $m$ backwards passes (one for each output). This observation is the root of preference for reverse-mode accumulation while training machine learning models (known as `backpropagation' in this special case); we differentiate a scalar-valued function with respect to a high-dimensional parameter vector (that is, $m=1, n \gg 1$).

\subsection{Federated Computations}\label{sec:fed_comp}

Our work focuses on taking derivatives of distributed functions, particularly functions whose computation spans clients and server in FL settings. In order to formalize this, we will use the concepts of \textbf{federated values} and \textbf{federated computations} as defined in the TensorFlow Federated framework~\citep{ingerman_ostrowski_2019}, and discussed in some detail by \citet{charles2022federated}. A \textbf{federated value} is some data hosted across a group of devices in a distributed system. Notably, it has a value and a \emph{placement} (that is, where is it located).
\footnote{The concept of placement in TensorFlow Federated is one of logical partition. It does not necessarily indicate the physical location of the device. While we use the placement and the physical location of data interdependently in our work, we note that placement can serve as a more general organizing mechanism (for example, by having multiple \CLIENTS placements that denote different sets of clients).}
We focus on two placements:
\begin{enumerate}
    \item \textbf{Server-placed values}. Conceptually, these are singleton values representing data directly accessibly by the server within a federated computation. While our work can be generalized to settings with multiple servers, we will assume throughout that there is a single server.
    \item \textbf{Client-placed values}. Conceptually, these are multiple values representing the data accessible by each client within a federated computation. We assume a form of semantic symmetry, so that if one client has a type of data (for example, a local set of model weights) then so too do all other clients participating in the federated computation. We model the collection of values across all participating clients (a set we denote by $C$) as a single federated value.
\end{enumerate}

When useful, we use $x\SERVER$ to denote a value of $x$ placed at the server, and $\{x_i \mid i \in C\}\CLIENTS$ to denote a client-placed federated value, where each participating client $i \in C$ has a corresponding value $x_i$. We often abbreviate the latter by $\{x_i\}\CLIENTS$ or even $x\CLIENTS$ (with the values $x_i$ implicit).

A \textbf{federated computation} is simply a function whose inputs and outputs are federated values. For example, a commonly-used federated computation is \fedbroadcast. This is simply the function that sends a server-placed value to all participating clients, ie:
\[
\fedbroadcast\left(x\SERVER\right) = x\CLIENTS
\]
where each client $i$ has the same value $x$. Another important federated computation is a \emph{federated sum}. This is simply the function given by:
\[
\fedsum\left(x\CLIENTS\right) = \left(\sum_{i\in C} x_i\right)\SERVER.
\]
Intuitively, this represents summing up values across all clients participating in a given round.

Given a (non-federated) function $g$ and placed values $x\SERVER$ and $z\CLIENTS$ (representing client values $\{z_i \mid i \in C\}$), we define
\begin{equation}\label{eq:server_comp}
g(x\SERVER) := g(x)\SERVER,
\end{equation}
\begin{equation}\label{eq:client_comp}
g(z\CLIENTS) := \{g(z_i) \mid i \in C\}\CLIENTS.
\end{equation}
In other words, we can form a federated computation from $g$ by applying it to local data (either at the server, or at each client).

FL algorithms are often designed with explicit \emph{data minimization} principles. This is typically done by preventing clients from directly sharing their own data with any other agent in the system. However, FL algorithms must be augmented with explicit privacy mechanisms if formal privacy guarantees are desired; see~\citep{aop} for detail on core mechanisms like differentially private aggregations and cryptographically secure aggregation (SecAgg, ~\citet{secagg}). We do not go into detail on these privacy-preserving mechanisms, except to note that many such mechanisms are often incompatible with server$\to$client communication that is client-dependent or client$\to$server communication that is not sum-based (such as median-based aggregation schemes).\footnote{For details on these restrictions, see \citep[Section 4]{aop}, especially the discussion of SecAgg and the shuffle model of differential privacy. See \citep{robust_agg} for an informative example of designing SecAgg-compatible robust aggregation methods.}
Thus, in privacy-sensitive settings, we generally want to restrict to federated computations that only use \fedbroadcast and \fedsum, in order to ensure compatibility with a wide array of privacy mechanisms. This precludes the use of things like client-dependent broadcast mechanisms (such as \fedselect~\citep{charles2022federated}) or median-based aggregation~\citep{yin2018byzantine}. While such communication primitives are not necessarily incompatible with formal privacy, they often require more specialized privacy mechanisms, and may depend strongly on the implementation~\citep{charles2022federated}.

\subsection{Motivating Questions}

We are now ready to state the primary questions motivating our work.
\begin{question}\label{question1}
How do we differentiate through federated computations?
\end{question}

There are naive answers to this that are impractical or lack compatibility with privacy-preserving technologies. For example, suppose we have a federated computation
\[
f(x\SERVER, y\CLIENTS) = z\SERVER
\]
and that we wish to compute the derivative of $z$ with respect to $x$. A non-private way to do this would be to implement the function $f$ in a single-process fashion; The clients could send their data to the server, and the server could then compute the derivative of $z$ with respect to $x$ using AD.
This approach clearly violates data-minimization principles of FL. Further, requiring a single process to compute and differentiate a function which accepts an input per-client could introduce scalability limitations; if the number of clients is large and the size of each constituent $y_i$ of $y\CLIENTS$ is also large, it may be infeasible for the server to directly compute $z$, let alone its derivative.

We would like to ensure that the server does not have direct access to client data, and that server-side computation can be implemented in a manner compatible with modern distributed systems. Therefore, we may wish to use privacy-preserving distributed technologies and existing distributed federated computation infrastructure to compute both $z$ and its derivative with respect to $x$. Thus, what we are actually interested in is the following:
\begin{question}\label{question2}
How do we differentiate through federated computations in a way that is both scalable and compatible with privacy-preserving technologies?
\end{question}

We give a partial answer to this in~\cref{sec:fed_diff}. In particular, we develop a framework, federated automatic differentiation (\fad) that enables us to compute derivatives for broad classes of federated computations. Moreover, we show that if a federated computation $f$ is compatible with certain privacy mechanisms (including differential privacy and secure aggregation), then we can use \fad to compute this kind of derivative of $f$ while maintaining compatibility with secure aggregation and differential privacy.\footnote{We will discuss several `modes` of \fad, analogous to forward- and reverse-mode in traditional AD. This claim is true for \emph{all} federated computations in forward-mode, and all computations which use summation to aggregate across clients in reverse-mode.}

While machine learning literature generally suggests that computing derivatives is important, it is not necessarily clear how one could put derivatives of federated computations to good use. Thus, after discussing \cref{question2}, we give a partial answer to the following question:
\begin{question}\label{question3}
How can we use federated automatic differentiation to design improved federated learning algorithms?
\end{question}

We discuss this question in \cref{sec:applying_fad}, to show that \fad can be used to develop self-tuning federated optimization algorithms. In \cref{sec:learned_server_opt} and \cref{sec:learned_weight}, we show that these self-tuning algorithms can yield improved performance over benchmark methods for federated optimization. We believe that \fad can enable significant work in the design of dynamic federated algorithms.

\section{Federated Automatic Differentiation}\label{sec:fed_diff}

In order to consider \cref{question2}, we begin by specializing the derivatives we compute. In the following, we differentiate with respect to server-placed parameters.\footnote{Our framework can be applied analogously to values that are constant across clients, but we restrict to server-placed values for simplicity of presentation. Note that we can model values that are constant across clients as the result of a call to \fedbroadcast.}
In practice, this restriction is likely to be minimal. Server-placed parameters are the traditional entry point by which a modeler or algorithm designer may specify behavior of a federated computation.

We illustrate our methods by considering a specific class of federated computations, though we note that \fad may be applied more generally. While relatively simple, this class illustrates our core approach for differentiating through federated computations.

\begin{class}\label{example:basic_fed_comp}
We assume that the server has some value, $x$, and each client $i$ has their local data $z_i$. We first consider federated computations of the form
\[
f(x\SERVER, z\CLIENTS) = y\SERVER
\]
that involve a single call to \fedbroadcast and \fedsum. For simplicity, we assume $f$ is defined as follows:
\begin{enumerate}\label{basic_fed_comp}
    \item Server holds $v_1\SERVER = x\SERVER$.
    \item Server computes $v_2\SERVER = f_1(v_1\SERVER)$.
    \item Clients receive $v_3\CLIENTS = \fedbroadcast(v_2\SERVER)$
    \item Clients compute $v_4\CLIENTS = f_2(v_3\CLIENTS, z\CLIENTS)$.
    \item Server receives $v_5\SERVER = \fedsum(v_4\CLIENTS)$.
    \item Server computes $y\SERVER = v_6\SERVER = f_3(v_5\SERVER)$.
\end{enumerate}
\end{class}

This class of functions includes the per-round logic of many notable algorithms, including \fedavg~\citep{mcmahan2017communication}, \fedprox~\citep{fedprox}, \perfedavg~\citep{perfedavg}, \fedopt~\citep{afo}, and many more.

\begin{figure}[t]
\caption{Federated computational graph for $y = f(x)$ as in \cref{example:basic_fed_comp}, where three clients participate. Here, $v_i$ represent intermediate values used in the computation. Each value has a placement (server or clients), and the clients have data $z_i$ which are not shared with one another or the server. Server$\to$client communication is done via \fedbroadcast, and client$\to$server communication is done via \fedsum.}
\includegraphics[width=0.95\linewidth]{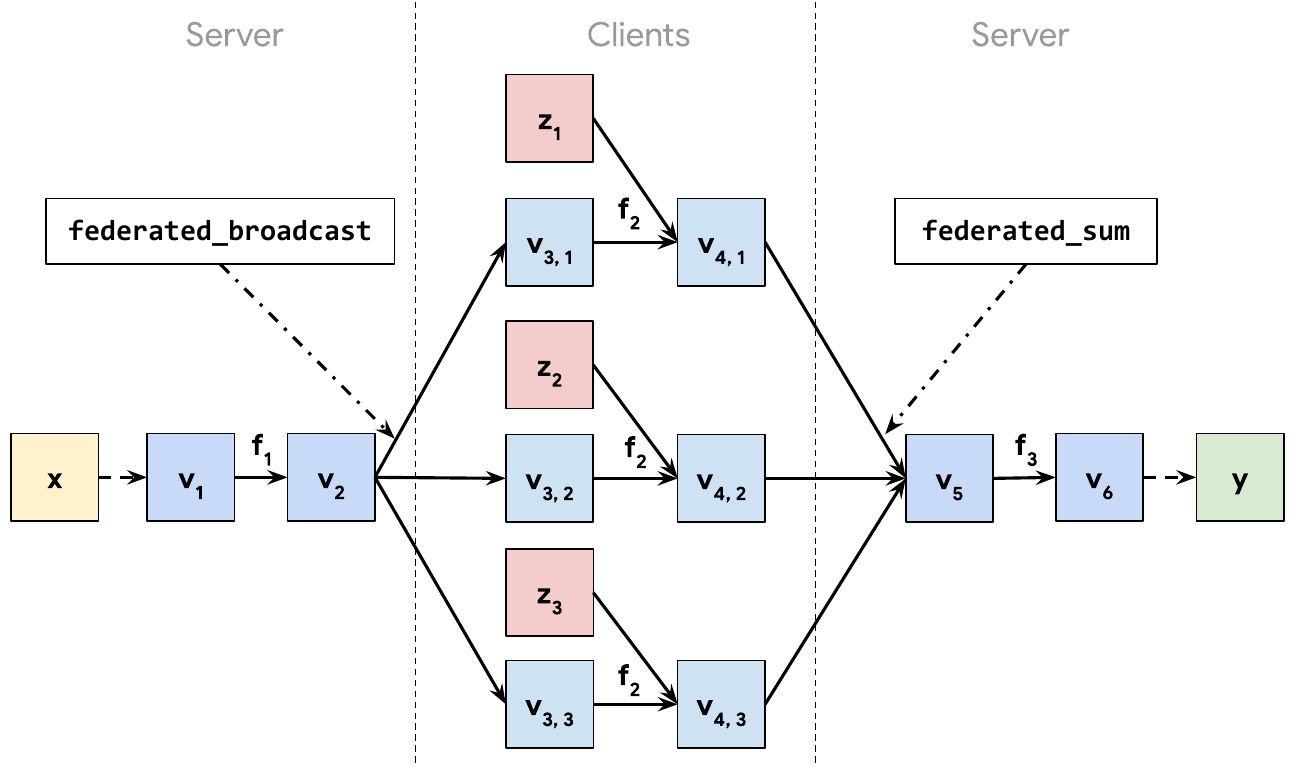}
\centering
\label{fig:fed_computation_graph_1}
\end{figure}

We wish to compute $\partial f/\partial x$ without sharing the local client data $z_i$. To understand how to do this, we first visualize $f$ as a \textbf{federated computational graph}. This is analogous to the computational graph in \cref{sec:auto_diff}, but where each value in the graph is a federated value (see \cref{sec:fed_comp}), where nodes are augmented with placements (server or clients). This is pictured in \cref{fig:fed_computation_graph_1}.

Ignoring data placement and assuming that the number of clients per round remains fixed, we could simply treat \cref{fig:fed_computation_graph_1} as a non-federated computational graph and use AD techniques. However, care must be taken when crossing communication boundaries so as not to violate data minimization principles of FL and to ensure compatibility with formal privacy mechanisms. Additionally, in many FL settings we would like to restrict the amount of communication incurred (and further may need to take into consideration asymmetry between download and upload bandwidth).

Below we present algorithms for \textbf{federated automatic differentiation} (\fad) that meet the above criteria. Notably, we give forward-mode, reverse-mode, and mixed-mode \fad algorithms, and show how they can be viewed as augmentations and transformations of federated communication patterns. Each of the approaches sketched below has distinct systems characteristics and requirements, as well as different amounts of communication incurred, as discussed in~\cref{ssec:fad_systems}.

\subsection{Forward-Mode \Fad}\label{sec:fwd_mode}

We begin with forward-mode \fad due to its simpler presentation. The crucial observation for applying forward-mode \fad to a federated computation $f$ is that it can be implemented using the same communication pattern as the computation of $f$. In particular, this means that we can use the same primitives (e.g. \fedbroadcast and \fedsum), but applied to forward-mode derivatives, in order to compute the desired derivative, without sharing the local data $z_i$.

We illustrate forward-mode \fad by stepping through differentiation of the function class described in~\cref{example:basic_fed_comp}. As in \cref{sec:auto_diff}, let $\dot{v}_i$ denote the Jacobian of $v_i$ with respect to $x$. We then have the following straightforward observations, consequence of the fact that derivatives commute with sums.\footnote{Here and in ~\cref{eq:reverse_mode_transform_1,eq:reverse_mode_transform_2} we rely on the fact that we specialize to differentiating with respect to a single, server-placed values. Differentiating with respect to a clients-placed value would instead result in multiple values (one per client). For example, one could argue that the correct derivative of a sum with respect to its clients-placed argument is the value $1\CLIENTS$.}
\begin{equation}\label{eq:forward_mode_transform_1}
    \dot{v}_{3}\CLIENTS = \fedbroadcast(\dot{v}_2\SERVER)
\end{equation}
\begin{equation}\label{eq:forward_mode_transform_2}
    \dot{v}_5\SERVER = \fedsum(\dot{v}_4\CLIENTS)
\end{equation}

According to the federated computational graph in \cref{fig:fed_computation_graph_1}, all that client $i$ needs to compute $\dot{v}_{4, i}$ are $z_i, v_{3, i}, \dot{v_{3, i}}$, which are available to the client, and $\dot{v}_{3, i}$, which they can receive via $\fedbroadcast(\dot{v}_2\SERVER)$ as per \eqref{eq:forward_mode_transform_1}. Thus, if the server computes $\dot{v}_2$, then the clients can compute $\dot{v}_{4, i}$. Similarly, for the server to compute $\dot{v}_6$, it requires $v_5$ and $\dot{v}_5$. This last quantity is simply $\fedsum(\dot{v}_{4}\CLIENTS)$ as per \eqref{eq:forward_mode_transform_2}.

This implies that forward-mode federated auto-differentiation can be done by following the same procedure as in \cref{example:basic_fed_comp}, but applying it to both the function values $v_j$ and derivatives $\dot{v}_j$. A side-by-side comparison of computing $f(x)$ versus computing $f(x)$ and $\partial f/\partial x$ via forward-mode \fad is given in \cref{table:forward_mode_fad}.  Our presentation here is drawn from \citep[Table 2]{autodiff_survey}, in part due to our admiration for the authors' wonderful exposition, and in part to demonstrate the conceptual similarities between forward-mode \fad and forward-mode AD.

\begin{table}[t]
  \centering
  \renewcommand{\arraystretch}{1.2}
  \caption{Forward mode \fad applied to $f$ in the class of functions described in \cref{example:basic_fed_comp}. We compare the procedure for computing $f(x)$ (left) with the augmented computation that also computes $\partial f/\partial x$ via forward mode \fad (right), with \fad-specific components marked {\color{blue}in blue}. Note that $x$ is a server-placed argument. We use $\dot{v}$ to denote $\partial v/\partial x$ (see \cref{sec:auto_diff}).}
  \label{table:forward_mode_fad}
  \begin{minipage}[c]{0.42\linewidth}
    {\footnotesize
    \begin{tabularx}{\textwidth}{p{0.1mm}p{70mm}X}
      \toprule
      \multicolumn{2}{l}{Evaluation of $y = f(x)$}\\
      \multirow{6}{2mm}{\begin{tikzpicture}\draw[->,>=triangle 60,thick](0,0)--(0,-3.6);\end{tikzpicture}} & Server receives input $v_1 = x$ \\
      & Server computes $v_2$\\
      & Clients receive: \\
      & \hspace{0.3cm} $v_3 = \fedbroadcast(v_2)$\\
      & Clients compute $v_4$\\
      & Server receives $v_5 = \fedsum(v_4)$\\
      & Server computes $v_6$ \\
      & Server outputs $y = v_6$\\
      \bottomrule
    \end{tabularx}}
  \end{minipage}
  \begin{minipage}[c]{0.57\linewidth}
    \setlength{\fboxsep}{0pt}\colorbox{gray!20}
    {\footnotesize
    \begin{tabularx}{\textwidth}{p{0.1mm}p{90mm}X}
      \toprule
      \multicolumn{2}{l}{Forward mode evaluation of $y = f(x)$ \color{blue}{and $\dot{y} = \partial f/\partial x$}}\\
      \multirow{6}{2mm}{\begin{tikzpicture}\draw[->,>=triangle 60,thick](0,0)--(0,-3.6);\end{tikzpicture}} & Server receives input $v_1 = x$ \\
      & Server computes $v_2$ \textcolor{blue}{and $\dot{v}_2$}\\
      & Clients receive: \\
      & \hspace{0.3cm} $v_3$, \textcolor{blue}{$\dot{v}_3$} = $\fedbroadcast(v_2, \textcolor{blue}{\dot{v}_2})$\\
      & Clients compute $v_4$ \textcolor{blue}{and $\dot{v}_{4}$}\\
      & Server receives $v_5, \textcolor{blue}{\dot{v}_5} = \fedsum(v_4, \textcolor{blue}{\dot{v}_4})$\\
      & Server computes $v_6$ \textcolor{blue}{and $\dot{v}_6$} \\
      & Server outputs $y = v_6$ \textcolor{blue}{and $\dot{y} = \dot{v}_6$}\\
      \bottomrule
    \end{tabularx}}
  \end{minipage}
\end{table}

Each of~\cref{eq:forward_mode_transform_1,eq:forward_mode_transform_2} illustrates an important principle: When performing \fad, derivatives must necessarily be communicated across device boundaries. Since communication bandwidth is often a critical limitation in federated systems, the size and shape of these derivatives may be crucial to the feasibility of computing the derivative of a particular federated computation. As we will see below, analogously to centralized AD, the size of the derivatives which flow across communication boundaries may be significantly altered by the mode in which we compute \fad.

\subsection{Reverse-Mode \Fad}\label{sec:rev_mode}

In centralized AD, particularly for machine learning applications, reverse-mode accumulation has obvious operational advantages to forward accumulation. Reverse-accumulation mode may also be defined in the federated setting, where, in direct analogy with centralized AD, it requires both re-addressing the same set of clients and reversing the communication patterns. 

Letting $\overline{v}_i$ denote $\partial y/ \partial v_j$, we have the following observations about \cref{fig:fed_computation_graph_1}:
\begin{equation}\label{eq:reverse_mode_transform_1}
    \overline{v}_2\SERVER = \fedsum(\overline{v}_3\CLIENTS)
\end{equation}
\begin{equation}\label{eq:reverse_mode_transform_2}
    \overline{v}_4\CLIENTS = \fedbroadcast(\overline{v}_5\SERVER)
\end{equation}

These mirror \eqref{eq:forward_mode_transform_1} and \eqref{eq:forward_mode_transform_2} for forward-mode \fad, but reverse the communication primitives involved. This gives us a straightforward way to perform reverse-mode \fad. We first do a forward pass over the computation of $f(x)$, following the procedure in \cref{example:basic_fed_comp}. Mirroring \cref{sec:auto_diff}, we use a backwards pass to compute $\overline{x}$. To do this, we reverse the arrows in \cref{fig:fed_computation_graph_1} and traverse the graph in reverse, starting at $y$. Whenever we encounter a value $v_j$, we use the available data from the forward-pass and previous nodes in the graph to compute $\overline{v}_j$. The only missing component is how to reverse the communication primitives used (\fedsum and \fedbroadcast). \cref{eq:reverse_mode_transform_1} and \cref{eq:reverse_mode_transform_2} imply that we simply swap the two. For example, the computation of $v_5\SERVER = \fedsum(v_4\CLIENTS)$ in the forward pass of \cref{example:basic_fed_comp} becomes $\overline{v}_4\CLIENTS = \fedbroadcast(\overline{v}_5\SERVER)$ in reverse-mode. Pseudo-code for this reverse-mode traversal is given in \cref{table:reverse_mode_fad}.

\begin{table}[t]
  \centering
  \renewcommand{\arraystretch}{1.2}
  \caption{An example of reverse-mode \fad when applied to the class of functions described in \cref{example:basic_fed_comp}. For comparison, we give a basic algorithm for computing $f(x)$ (left), as well as a backwards computation that also computes $\partial f/\partial x$ via reverse-mode \fad (right). Note that the reverse-mode computations depend on access to the previously computed values in the forward pass. We use $\overline{v}$ to denote $\partial y/\partial v$ (see \cref{sec:auto_diff}).}
  \label{table:reverse_mode_fad}
  \begin{minipage}[c]{0.49\textwidth}
    {\footnotesize
    \begin{tabularx}{\textwidth}{p{0.1mm}p{70mm}X}
      \toprule
      \multicolumn{2}{l}{Evaluation of $y = f(x)$}\\
      \multirow{6}{2mm}{\begin{tikzpicture}\draw[->,>=triangle 60,thick](0,0)--(0,-3.1);\end{tikzpicture}} & Server receives input $v_1 = x$ \\
      & Server computes $v_2$\\
      & Clients receive $v_3 = \fedbroadcast(v_2)$\\
      & Clients compute $v_4$\\
      & Server receives $v_5 = \fedsum(v_4)$\\
      & Server computes $v_6$ \\
      & Server outputs $y = v_6$\\
      \bottomrule
    \end{tabularx}}
  \end{minipage}
  \begin{minipage}[c]{0.49\textwidth}
    \setlength{\fboxsep}{0pt}\colorbox{gray!20}
    {\footnotesize
    \begin{tabularx}{\textwidth}{p{0.1mm}p{70mm}X}
      \toprule
      \multicolumn{2}{l}{Reverse mode evaluation of $\overline{x} = \partial f/\partial x$}\\
      \multirow{6}{2mm}{\begin{tikzpicture}\draw[<-,>=triangle 60,thick](0,0)--(0,-3.1);\end{tikzpicture}} & Server outputs $\overline{x} = \overline{v}_1$\\
      & Server computes $\overline{v}_1$\\
      & Server receives $\overline{v}_2 = \fedsum(\overline{v}_3)$\\
      & Clients compute $\overline{v}_3$\\
      & Clients receive $\overline{v}_4 = \fedbroadcast(\overline{v}_5)$ \\
      & Server computes $\overline{v}_5$ \\
      & Server receives input $\overline{v}_6 = \overline{y}$\\
      \bottomrule
    \end{tabularx}}
  \end{minipage}
\end{table}

Notably, reverse-mode \fad only uses \fedsum and \fedbroadcast to perform communication, and does not involve directly sharing client data. It is therefore compatible with the same privacy-preserving mechanisms as forward-mode \fad. It also shares some of the advantages of reverse-mode centralized AD, as discussed in~\cref{ssec:fad_systems}.

\subsection{Mixed-Mode \Fad}\label{sec:mix_mode}

In non-federated settings, forward-mode and reverse-mode AD can both be viewed as traversals of a computational graph, where the derivative is accumulated as we traverse. Notably, these two modes are in some sense ``extreme'': we either start at a root of the corresponding DAG and only move forward, or start at a leaf and only move backwards. However, as long as we correctly apply the chain rule, we have freedom to traverse the graph in all manner of ways (e.g. moving forward across some edges, and backwards across others). In fact, such ``mixed-mode'' accumulation of derivatives can result in a smaller number of required arithmetic operations (though finding the minimum number of arithmetic operations required is NP-complete)~\citep{oja}.

Mixed-mode differentiation is attractive in the federated setting, as different devices may face different limits on computation and communication, and there may be asymmetry in cost between different types of communication (e.g. up-link vs down-link). In general, we often wish to minimize communication to and from clients, and minimize the total amount of computation on the client. We may also wish to remove the systems issues which hinder implementation of reverse-mode \fad, where clients dropping out between the forward-pass and reverse-pass can lead to bias in the gradient computation.

We can do this by decoupling the overall traversal of the federated computational graph in \cref{fig:fed_computation_graph_1} from the local graph traversal done by the server or clients. Specifically, we can use the chain rule to move forward through the communication primitives used (the \fedbroadcast and \fedsum calls, in that order) but use reverse-mode locally at each device (for example, we can use reverse-mode for the part of the computation that occurs solely at the clients). The resulting procedure is given in \cref{table:mixed_mode_fad}.

\begin{table}[t]
  \centering
  \renewcommand{\arraystretch}{1.2}
  \caption{Mixed-mode \fad applied to $f$ in the class of functions described in \cref{example:basic_fed_comp}. We compare the procedure for computing $f(x)$ (left) with the mixed-mode \fad computations (right). While the communication pattern allows mixed-mode \fad to be done in tandem with the forward-pass, derivatives are computed locally by the server or clients using reverse-mode (not federated) AD.}
  \label{table:mixed_mode_fad}
  \begin{minipage}[c]{0.48\textwidth}
    {\footnotesize
    \begin{tabularx}{\textwidth}{p{0.1mm}p{65mm}X}
      \toprule
      \multicolumn{2}{l}{Evaluation of $y = f(x)$}\\
      \multirow{6}{2mm}{\begin{tikzpicture}\draw[->,>=triangle 60,thick](0,0)--(0,-5.2);\end{tikzpicture}} & Server receives input $v_1 = x$ \\
      & Server computes $v_2$\\\cmidrule{2-2}
      & Clients receive $v_3 = \fedbroadcast(v_2)$\\
      & Clients compute $v_4$\\\cmidrule{2-2}
      & \multirow{2}{90mm}{Server receives $v_5 = \fedsum(v_4)$}\\
      & \\\cmidrule{2-2}
      & \multirow{2}{90mm}{Server computes $v_6$} \\
      & \\\cmidrule{2-2}
      & \multirow{2}{90mm}{Server outputs $y = v_6$}\\
      & \\
      \bottomrule
    \end{tabularx}}
  \end{minipage}
  \begin{minipage}[c]{0.51\textwidth}
    \setlength{\fboxsep}{0pt}\colorbox{gray!20}
    {\footnotesize
    \begin{tabularx}{\textwidth}{p{0.1mm}p{0.1mm}p{63mm}X}
      \toprule
      \multicolumn{3}{l}{Mixed-mode evaluation of $\partial y/\partial x$}\\
      \multirow{6}{2mm}{\begin{tikzpicture}\draw[->,>=triangle 60,thick](0,0)--(0,-5.2);\end{tikzpicture}} & \multirow{2}{2mm}{\begin{tikzpicture}\draw[<-,>=triangle 60,thin](0.1,0)--(0.1,-0.8);\end{tikzpicture}} & \multirow{2}{90mm}{Server computes $\dfrac{\partial v_2}{\partial v_1}$ via reverse-mode AD} \\\
      & & \\\cmidrule{3-3}
      & \multirow{2}{2mm}{\begin{tikzpicture}\draw[<-,>=triangle 60,thin](0.1,-1.1)--(0.1,-1.9);\end{tikzpicture}} & \multirow{2}{90mm}{Clients compute $\dfrac{\partial v_4}{\partial v_3}$ via reverse-mode AD} \\
      & & \\\cmidrule{3-3}
      & & \multirow{2}{70mm}{Server receives $\dfrac{\partial v_5}{\partial v_2} = \fedsum\left(\dfrac{\partial v_4}{\partial v_3}\right)$}\\
      & & \\\cmidrule{3-3}
      & \multirow{2}{2mm}{\begin{tikzpicture}\draw[<-,>=triangle 60,thin](0.1,-2.0)--(0.1,-2.8);\end{tikzpicture}} & \multirow{2}{90mm}{Server computes $\dfrac{\partial v_6}{\partial v_5}$ via reverse-mode AD}\\
      & & \\\cmidrule{3-3}
      & & \multirow{2}{90mm}{Server outputs $\dfrac{\partial y}{\partial x} = \dfrac{\partial v_6}{\partial v_1} = \dfrac{\partial v_6}{\partial v_5}\dfrac{\partial v_5}{\partial v_2}\dfrac{\partial v_2}{\partial v_1}$}\\
      & & \\
      \bottomrule
    \end{tabularx}}
  \end{minipage}
\end{table}

In comparison to forward-mode \fad, we see mixed-mode does not require sending (potentially large) Jacobian matrices to the clients, and allows clients to use reverse-mode AD locally. In comparison to reverse-mode \fad, we again see that no extra server$\to$client communication is necessary, and that the derivative computations can be done in tandem with the forward pass.

Just as forward- and reverse-mode \fad can be extended to larger classes of computations, so too can mixed-mode \fad, using similar principles. In general, mixed-mode \fad can be applied to any computation whose communication is performed via \fedbroadcast and \fedsum, and the derivative computation itself only requires \fedsum. As a result, mixed-mode \fad is also compatible with formal privacy mechanisms. Finally, alternate specifications of mixed-mode \fad are possible, which may be more or less desirable based on the particulars of the implementing system and the computation being differentiated.

\subsection{System Considerations}\label{ssec:fad_systems}

We concretize an example from~\cref{example:basic_fed_comp} and walk through differentiation in the three modes discussed above in order to highlight the systems and communication considerations at play when choosing a \fad mode. 

Suppose that the function $f_1$ of~\cref{example:basic_fed_comp} (which runs on the server before the model broadcast) performs a \emph{distillation} of the incoming server-side model $x$. That is, suppose $x$ represents a vector of dimensionality \largemodel~and $f_1(x)$ computes a vector $u$ of dimensionality \smallmodel, where $\smallmodel \ll \largemodel$. The function $f_2$ computes the loss of a client on the distilled model, and the function $f_3$ is simply the identity function. Thus, the remainder of our computation simply broadcasts and evaluates the distilled model $u$ across the clients before computing an average of the losses on the server, which we denote by $y$. This computation broadcasts $\smallmodel$ floats and aggregates a single float from each client.

Now, suppose we wish to differentiate $y$ with respect to $x$. In \textbf{forward-mode}, we compute and communicate $\frac{du}{dx}$, which in our case (assuming matrices act on the left) is a matrix $\textbf{A}$ of shape $\smallmodel \times \largemodel$. We broadcast this (potentially quite large) matrix to the clients, which locally compute the derivative of their local losses with respect to their incoming models, which may be represented as matrices $\textbf{B}$ of shape $1 \times \smallmodel$. Clients must then perform the multiplication $\textbf{B}\textbf{A}$, resulting in each client owning a matrix of shape $1 \times \largemodel$, which will then be averaged. Note that the floats broadcasted and aggregated in forward-mode are both inflated by a multiplicative factor of $\largemodel$, the dimensionality of the value with respect to which we differentiate.

In \textbf{reverse-mode}, at the end of the `forward pass' of the computation, the server broadcasts a \emph{scalar} value to each client. Clients then compute $\frac{dy}{du}$, of shape $1 \times \smallmodel$. This value is then summed by the server and leveraged to compute $\frac{dy}{dx}$, a matrix of shape $1 \times \largemodel$, via local reverse-accumulation mode---in particular, without ever materializing the matrix $\textbf{A}$ above.

In \textbf{mixed-mode}, we might pin all the matrix multiplies to the server, asking clients to compute the $1 \times \smallmodel$ matrix representing derivative of local loss with respect to the broadcast model $u$. The server maintains (rather than broadcasts) the matrix state $\textbf{A}$ of shape $\smallmodel \times \largemodel$ (or the capacity to compute matrix-vector products with respect to $\textbf{A}$), and performs a post-aggregation chain-rule multiplication with the aggregated client-side derivatives of shape $1 \times \smallmodel$ to compute the derivative $\frac{dy}{dx}$ of shape $1 \times \largemodel$.

Several interesting features of these various implementations come immediately to light. Each of these modes has distinct computation and communication properties. Mixed mode actually communicates the \emph{least}, though requires materializing the large matrix $\textbf{A}$ (or the capacity to peform ex-post-facto vector-Jacobian products, effectively splitting up the implementation of the chain rule). Still, it can be implemented without assuming stable client connections. Reverse mode, despite its appealing communication and computation properties, requires this assumption, for the same reason that intermediate values must be stored (or recomputed) while performing backpropagation in centralized ML. Many existing cross-device FL systems are MapReduce-like in their structure~\citep{fl_at_scale}, \emph{designed without the ability to re-address the same cohort of clients}. This means that such systems are incompatible with the kind of reverse-mode \fad computations in \cref{table:reverse_mode_fad}. Finally, though forward-mode \emph{significantly} increases the communication costs of both upload and download, in some sense this is an artifact of the chosen example: forward-mode differentiating a vector-valued function with respect to a scalar parameter would show forward-mode \fad to be preferable to reverse-mode.

Observations about the systems characteristics of the various modes of \fad may be generalized to yield statements about the possibility of communication-efficient implementations. Preference for one implementation over another is potentially dependent on both the inputs and outputs of the function being differentiated as well as the cost model of the targeted federated system (e.g., as illustrated above, relative cost between upload and download may affect choice of forward versus reverse-mode).

 \section{Applying Federated Automatic Differentiation to Federated Learning}\label{sec:applying_fad}

We now turn our attention to applying \fad to FL tasks, particularly federated optimization. Throughout, we do not focus on creating ``state-of-the-art'' methods for federated optimization. Instead, we intend to demonstrate that \fad can enable more expressive algorithms in which components of the federated computation are learned throughout training, and to good empirical effect.

While one can apply \fad to a wide array of federated computations, for concreteness we will focus on applying it to learn portions of the \fedopt algorithmic framework~\citep{afo}. \fedopt is an empirically successful and widely adopted federated optimization paradigm that combines client-level and server-level optimization.

\subsection{\fedopt Framework}

\fedopt works as follows: At each round $t$, the server has a model $x_t$. This model is broadcast to some set $C_t$ of participating clients (via \fedbroadcast). Each client $i \in C_t$ computes an updated model $x_{t, i}$ using some procedure \clientupdate (typically some number of first-order optimization steps) with the client's local data $z_i$. The client then computes $\Delta_{t, i} = x_t - x_{t, i}$. The server receives the average $\Delta_t$ of these $\Delta_{t, i}$ from the clients, and uses this to update the server model $x_t$ via some procedure \serverupdate (typically a first-order optimization step).

{
\begin{algorithm}[t]
\caption{\fedopt}
\label{alg:fedopt}
\begin{algorithmic}[1]
	    \State Input: server weights $x_0\SERVER$, client data $\{z_i\}\CLIENTS$, client weights $\{\rho_i\}\CLIENTS$
	    \For{$t = 0, \cdots, T-1$}
    	    \State Sample a cohort $C_t$ of participating clients.
    	    \State Clients receive $\fedbroadcast(x_t\SERVER)$
    	    \For{each client $i \in C_t$ \textbf{in parallel}}
        	    \State $x_{t, i} = \clientupdate(x_t, z_i)$
        	    \State $\Delta_{t, i} = x_{t} - x_{t, i}$
    	    \EndFor
    	    \State Server receives $\Delta_t = \fedmean\left(\{\Delta_{t, i}\}\CLIENTS, \{\rho_i\}\CLIENTS\right)$
    	    \State Server computes $x_{t+1} = \serverupdate(x_t, \Delta_t)$
		\EndFor
\end{algorithmic}
\end{algorithm}
}

Pseudo-code for \fedopt is given in \cref{alg:fedopt}. In this algorithm, the server receives a mean of the client ``model deltas'' $\Delta_{t, i}$ weighted (respectively) by scalars $\rho_i$. We denote this server-placed federated value by  $\fedmean(\{\Delta_{t, i}\}\CLIENTS, \{\rho_i\}\CLIENTS)$.
Note that this operation can be reduced to \fedsum since
\begin{equation*}
\fedmean(\{\Delta_{t, i}\}\CLIENTS, \{\rho_i\}\CLIENTS) = \dfrac{\fedsum(\{\rho_i\Delta_{t, i}\}\CLIENTS)}{\fedsum(\{\rho_i\}\CLIENTS)}.
\end{equation*}
In particular, this means that the communication in \fedopt only involves \fedbroadcast and \fedsum, and is therefore compatible with privacy-preserving technologies (e.g. SecAgg and differential privacy).

The client weights $\rho_i$ are typically constants that are easily computed by each client. Two common weighting schemes are \emph{example weighting}, where each client's weight is the number of examples they hold (so that $\rho_i = |z_i|$), and \emph{uniform weighting}, where each client's weight is the same (so that $\rho_i = 1$). Example weighting was proposed by \citet{mcmahan2017communication}, and shown to be effective from an optimization standpoint by \citet{fedprox}. However, uniform weighting is necessary for certain privacy mechanisms like differentially private averages, which are often unweighted to ease control of sensitivity.

\begin{figure}[t]
\caption{Federated computational graph for a single round of \fedopt.}
\includegraphics[width=0.9\linewidth]{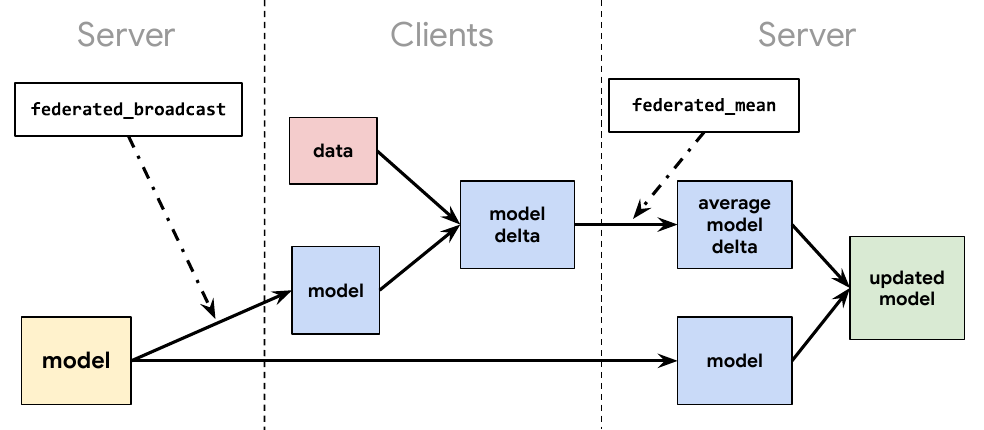}
\centering
\label{fig:fedopt}
\end{figure}

For reference, we give a federated computational graph representing one round of \fedopt in \cref{fig:fedopt}. We only picture a single client, but conceptually the same client operations are being done in parallel on some set of participating clients (as in \cref{fig:fed_computation_graph_1}). We will return to \cref{fig:fedopt} in later sections, in order to discuss applications of \fad to \fedopt.

\subsection{Federated Hypergradient Descent}

As discussed in \cref{sec:intro}, we would like to enable FL algorithms that dynamically adjust their structure as they train. We now detail how to use \fad to learn aspects of \fedopt in tandem with execution of \fad by adjusting existing components of \fedopt. We do this by applying \emph{federated hypergradient descent} to \cref{alg:fedopt}. The primary obstacle to this is that individual clients can only compute hypergradients with respect to their own loss function, while the server cannot directly compute any hypergradients (as it does not necessarily possess any data). Thus, in the absence of \fad, it is not immediately clear how to compute derivatives of the form $\partial L(x)/\partial \alpha$ for some global loss function $L$ at a model $x$ with respect to some server hyperparameter $\alpha$.

This roadblock was noted by prior works on federated hyperparameter tuning. While \citet{fed_hparam_tuning} use a sophisticated weight-sharing mechanism to learn client hyperparameters, they learn server hyperparameters by training multiple times with different hyperparameter settings and eliminating parameters that are not doing well (e.g. via successive halving). While effective, this procedure relies on training multiple models. More promisingly, \citet{fed_hypergrad_descent} applies hypergradient descent to some hyperparameters of \fedopt. However, this work only considers certain client hyperparameters (notably, only hyperparameters of \clientupdate), and relies on hand-derived gradients (which also use various approximations).

\begin{figure}[t]
\caption{Federated computational graph used to compute server hypergradients in \fedopt. All server$\to$client communication is done via \fedbroadcast, while all client$\to$server is done via \fedmean. This computation produces some updated model and an estimate of the loss of that model.}
\includegraphics[width=0.98\linewidth]{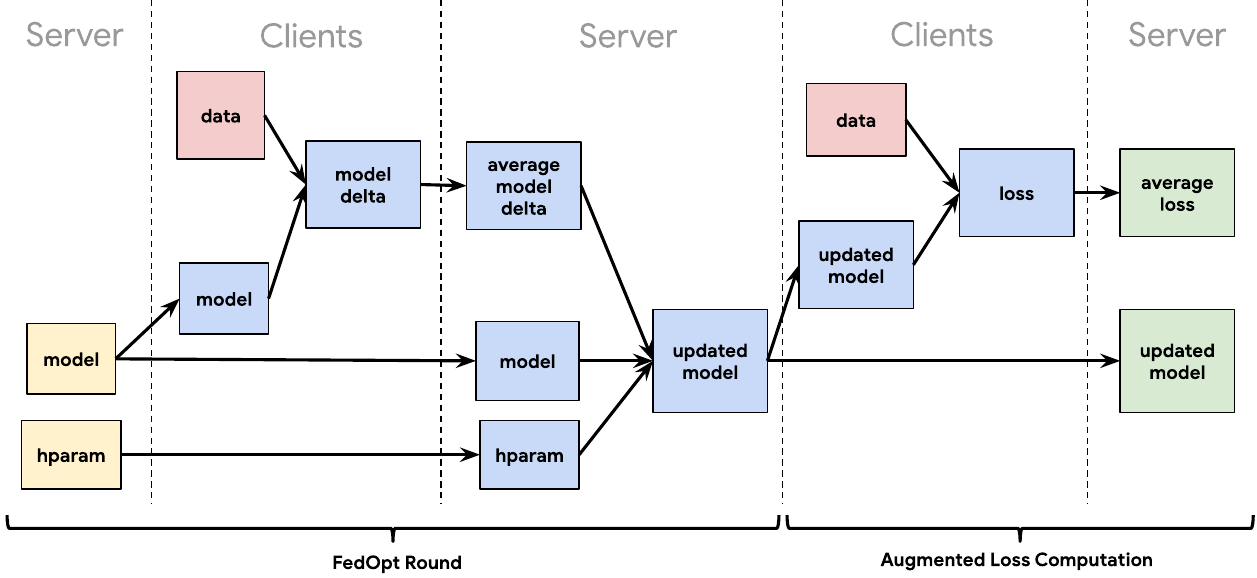}
\centering
\label{fig:server_hparam_serial}
\end{figure}

\Fad gives us a straightforward way to compute $\partial L/\partial \alpha$ for hyperparameters without the need to derive explicit derivative formulas. In order to apply \fad, we augment each round of \fedopt (\cref{alg:fedopt}) with a computation in which 1) the server broadcasts its updated model to a set of clients, 2) each client computes their associated loss, and 3) the server receives the average of client loss. By using \fad to differentiate through this, we can arrive at the desired derivative. This can be done via any of the modes discussed in \cref{sec:fed_diff}.

\cref{fig:server_hparam_serial} gives the associated federated computational graph to which we apply \fad, specialized to a server-side hyperparameter. This federated computational graph has two components: A single round of \fedopt (essentially the same graph as \cref{fig:fedopt}), followed by a federated loss computation. This produces some loss measurement $L(x)$. By applying \fad (forward-mode, reverse-mode, or mixed-mode), we can differentiate through \cref{fig:server_hparam_serial} to get $\partial L/\partial \alpha$. Once we have this derivative, we can then perform a gradient-descent step to update $\alpha$. High-level pseudo-code for this entire procedure is given in \cref{alg:fedopt_hparam}. To simplify notation, we let  $\fedoptround(x, \alpha, \{z_i\}_{i \in C})$ denote the model update produced by a single round of \fedopt, using some hyperparameter $\alpha$, with client data $\{z_i\}_{i \in C}$ for a set $C$ of participating clients. Note that other first-order optimization methods could be used to update the hyperparameter $\alpha$.

{
\begin{algorithm}[t]
\caption{\fedopt with hypergradient descent.}
\label{alg:fedopt_hparam}
\begin{algorithmic}[1]
	    \State Input: $x_0\SERVER$, $\alpha_0$, $\eta\SERVER$, $\{z_i\}\CLIENTS$
	    \For{$t = 0, \cdots, T-1$}
    	    \State Sample a cohort $C_t$ of participating clients.
    	    \State Server receives $x_{t+1} = \fedoptround(x_t, \alpha_t, \{z_i\}_{i \in C_t})$ and $\partial f(x_{t+1})/\partial \alpha_t$ via \fad, as in \cref{fig:server_hparam_serial}.
    	    \State Server computes $\alpha_{t+1} = \alpha_t - \eta (\partial f(x_{t+1})/\partial \alpha_t)$.
		\EndFor
\end{algorithmic}
\end{algorithm}
}

\begin{figure}[t]
\caption{Federated computational graph used to compute hypergradients of server hyperparameters in \fedopt. All server$\to$client communication is done via \fedbroadcast, while all client$\to$server is done via \fedmean. This computation produces some updated model and an estimate of the loss of that model. In contrast to \cref{fig:server_hparam_serial}, model training and loss computation (for the purposes of computing hypergradients) are done in parallel, potentially across different sets of clients. Note that applying \fad to compute the derivative of the average loss with respect to the hyperparameter requires chaining two of these graphs together (in order to create a path from ``hparam'' to ``average loss''), though this can be performed by the server after the fact leveraging only a direct application of \fad to the graph presented here.}
\includegraphics[width=0.75\linewidth]{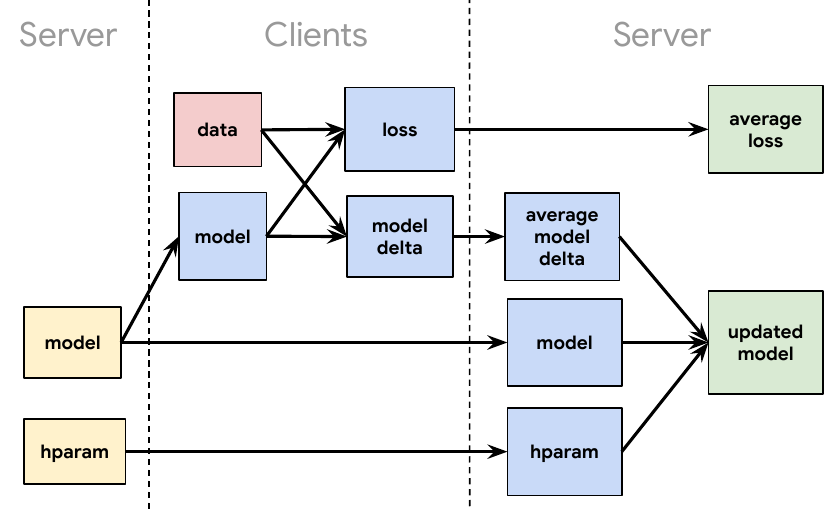}
\centering
\label{fig:server_hparam_parallel}
\end{figure}

We make a few important observations about \cref{fig:server_hparam_serial} and \fad.
\begin{enumerate}
    \item The augmented loss computation only provides a stochastic estimate of the true loss $f(x)$, and therefore a stochastic hypergradient $\partial f(x)/\partial \alpha$. Generally, this will depend on which clients participate during the augmented loss computation. However, the set of clients which participate in the augmented loss computation need not be the same as the clients used for the \fedopt model update. This is useful in FL settings where clients have limited availability, and may drop out between computations.
    \item Due to the serial nature of the \fedopt round and augmented loss computation, we effectively double the number of communication rounds needed (relative to \fedopt). This can be important in FL settings (especially cross-device settings) where synchronization costs are high~\citep{fl_at_scale}. In order to avoid this, we can parallelize the \fedopt computation and augmented loss computation into a single communication round, as pictured in \cref{fig:server_hparam_parallel}.
\end{enumerate}

\section{Learned Server Optimization}\label{sec:learned_server_opt}

We now present an empirical exploration of federated hypergradient descent as applied to server optimization hyperparameters. We instantiate the \fedopt framework with the \fedavgm algorithm~\citep{fedavgm}: namely, \cref{alg:fedopt} where \clientupdate is $E$ epochs of SGD (as in \fedavg), and where \serverupdate is a single step of SGD with momentum. We differentiate with respect to learning rate and momentum as detailed in \cref{sec:applying_fad} to pursue twin questions. First: can following gradients make \fedavgm easier to tune? Second: can following gradients recover and surpass performance of fixed hyperparameter settings for \fedavgm?

\subsection{Experimental Setup}

We tested our algorithm across five tasks adapted from \citet{afo}. We used four data sets: CIFAR-100~\citep{krizhevsky2009learning}, EMNIST~\citep{cohen2017emnist}, Shakespeare~\citep{mcmahan2017communication}, and Stack Overflow~\citep{stackoverflow}.
For EMNIST, we train a relatively small CNN for character recognition. For CIFAR-100, we train a modified ResNet-18 model, replacing BatchNorm layers with GroupNorm~\citep{groupnorm_for_fl}.
For Shakespeare, we trained an RNN for next-character-prediction.
For Stack Overflow, we performed tag prediction using a logistic regression on bag-of-words vector (SO TP) and trained an RNN to do next-word-prediction (SO NWP).

Throughout, we compare the ``accuracy'' of the learned model on the test split of each of the data sets above. This refers to the standard multi-class prediction accuracy for the CIFAR-100, EMNIST, Shakespeare, and Stack Overflow NWP tasks. For Stack Overflow TP, the ``accuracy'' refers to recall@5, following the convention in \cite{afo}.

We use $E = 1$ epochs of mini-batch SGD in \clientupdate throughout, and use the same batch sizes for each task as in \citep{large_cohort}. We set the per-client weights $p_k$ to the number of examples in each client's data set (example-weighting). We sample 50 clients uniformly at random at each communication round. We use the same number for any derivative computations performed via \fad, though these need not be coupled.

We compare various initialization strategies for the client learning rate, server learning rate, and server momentum. In each case, we either set it to some default value, or a random value. The random initializations use the same sampling strategy across tasks, detailed in~\cref{app:server_opt_experimental_details}. The default client learning rates are taken from the tuned learning rates in \citep{afo}. The default server learning rate is set to $1.0$, and the default momentum value is set to 0.9. We then compare random/default client learning rates and random/default server parameters, for a total of four initialization strategies.

We compute derivatives of the loss with respect to the server learning rate and momentum using a limited implementation of mixed-mode \fad from \cref{sec:fed_diff}, as detailed in~\cref{app:server_opt_experimental_details}. We use the ``parallelized'' approach to \cref{alg:fedopt_hparam} presented in \cref{fig:server_hparam_parallel}. Crucially, we \emph{do not tune the hypergradient descent optimizer} we used to adjust the server momentum and learning rate parameters over time. We performed some initial small tuning on the EMNIST task, and simply reused this setting for all the rest. Throughout, our hypergradient descent optimizer is SGD with a learning rate of 0.01.

\subsection{Results}

We perform 50 random trials for each of the 4 default/random combinations and task. Across all initializations, learned learning rate and momentum values essentially always outperformed their fixed counterparts on all metrics we evaluated. We consider one of our initialization settings to be particularly interesting, and we discuss these results in detail: initialization to some `default' server optimizer settings (learning rate of 1.0 and momentum of 0.9, a standard baseline setup in FL), and random client learning rates. Note that we see qualitatively similar results for the other three combinations of default/random, see \cref{app:server_opt_experimental_details}.

The default server / random client setting is intended to replicate a `first-pass' hyperparameter tuning attempt for a FL practitioner. Without adaptive optimizers, FL with server learning rate of 1.0 is usually a reasonable default; indeed, the original formulation of \fedavg assumes a server learning rate of 1.0 when translated to the bi-level optimization setting of~\citet{afo}. The momentum parameter of 0.9 served as a well-tuned default in~\citet{afo}, though usually some small improvement can be found for a given task by tuning the momentum value. Little guidance exists, however, for choosing a client learning rate, and the optimal client learning rate for a federated tasks often differs substantially from the optimal learning rate for the same task in the centralized setting. In practice, this is effectively always swept over. This client learning rate sweep comes at considerable computational cost, particularly from the overall system health perspective, where tasks may compete for resources at training time~\citep{fl_at_scale}.

\begin{table}[t!]
\centering
    \begin{tabular}{|c|c|c|c|c|c|}
    \hline
    Learned &  SO NWP & Shakespeare & EMNIST & CIFAR100 & SO Tag\\
    \hline
    True & \textbf{25.7} & \textbf{57.3} & \textbf{87.1} & \textbf{45.6} & \textbf{35.8} \\
    False  & 25.0 & 56.8 & 87.0 & 43.2 & 32.2 \\
    \hline
\end{tabular}
\caption{Maximum test accuracy (\%) across 50 trials when using \fad to learn the learning rate and momentum parameters of server-side SGD. We initialize with learning rate $1.0$ and momentum $0.9$, and a random client learning rate.}
\label{tab:random_server_default_client}
\end{table}

\begin{figure}
\begin{subfigure}{.5\textwidth}
  \centering
  \includegraphics[width=.9\linewidth]{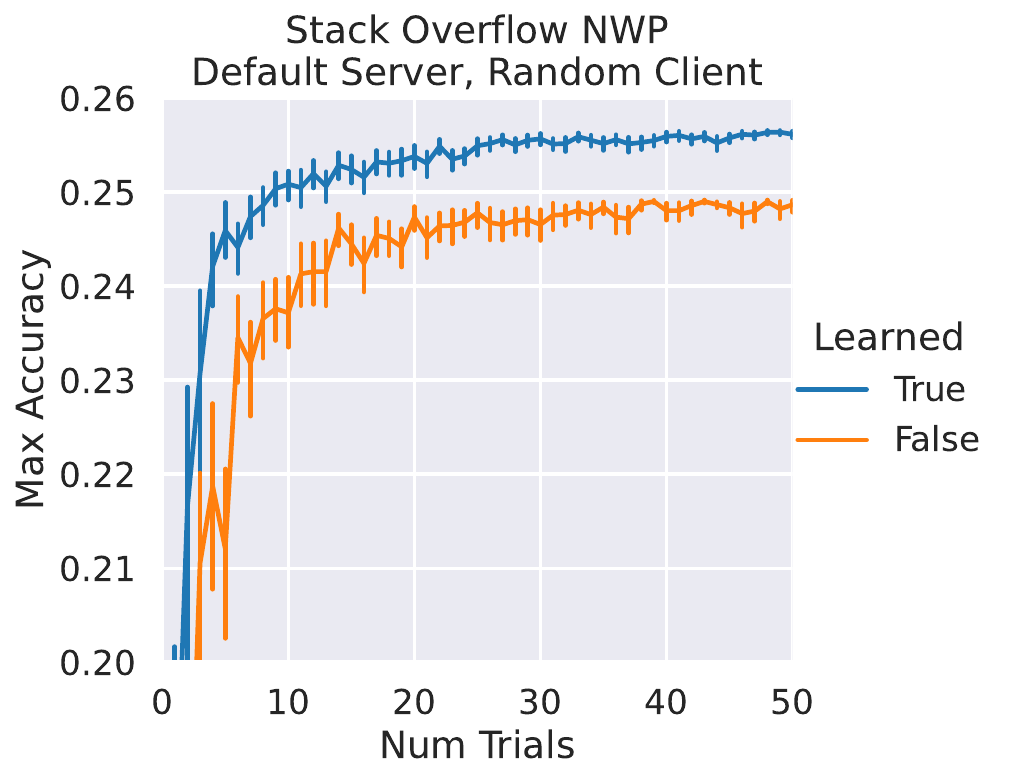}
  \label{fig:so_nwp}
\end{subfigure}%
\begin{subfigure}{.5\textwidth}
  \centering
  \includegraphics[width=.9\linewidth]{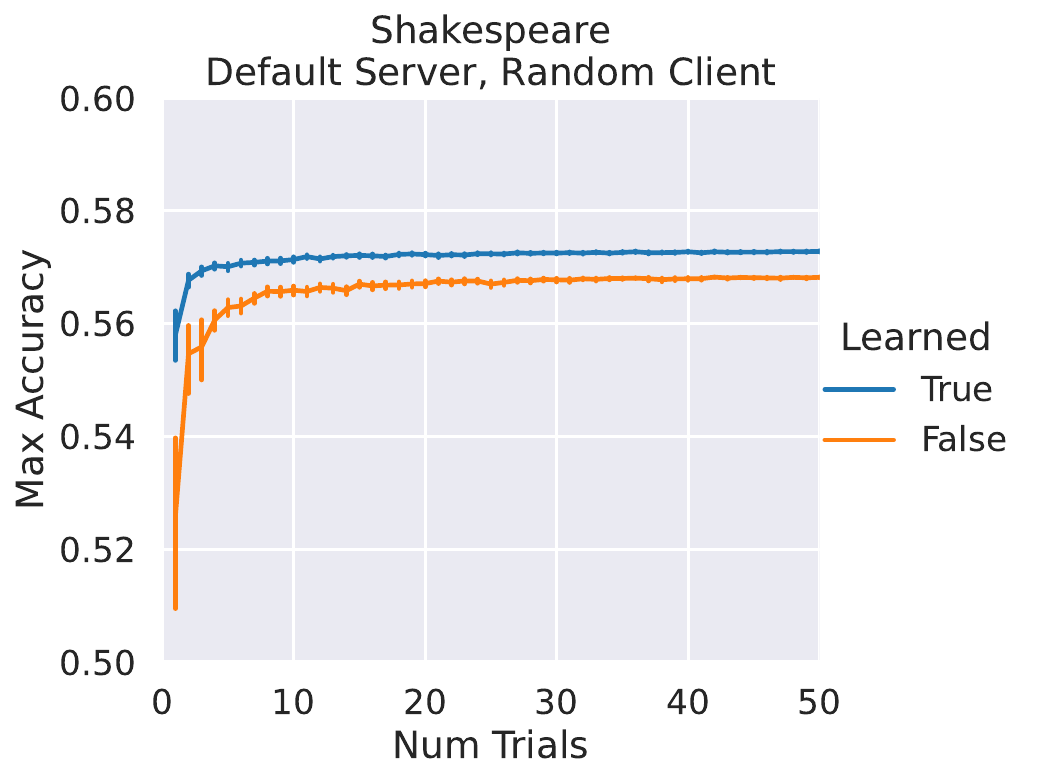}
  \label{fig:shakespeare}
\end{subfigure}
\begin{subfigure}{.5\textwidth}
  \centering
    \includegraphics[width=.9\linewidth]{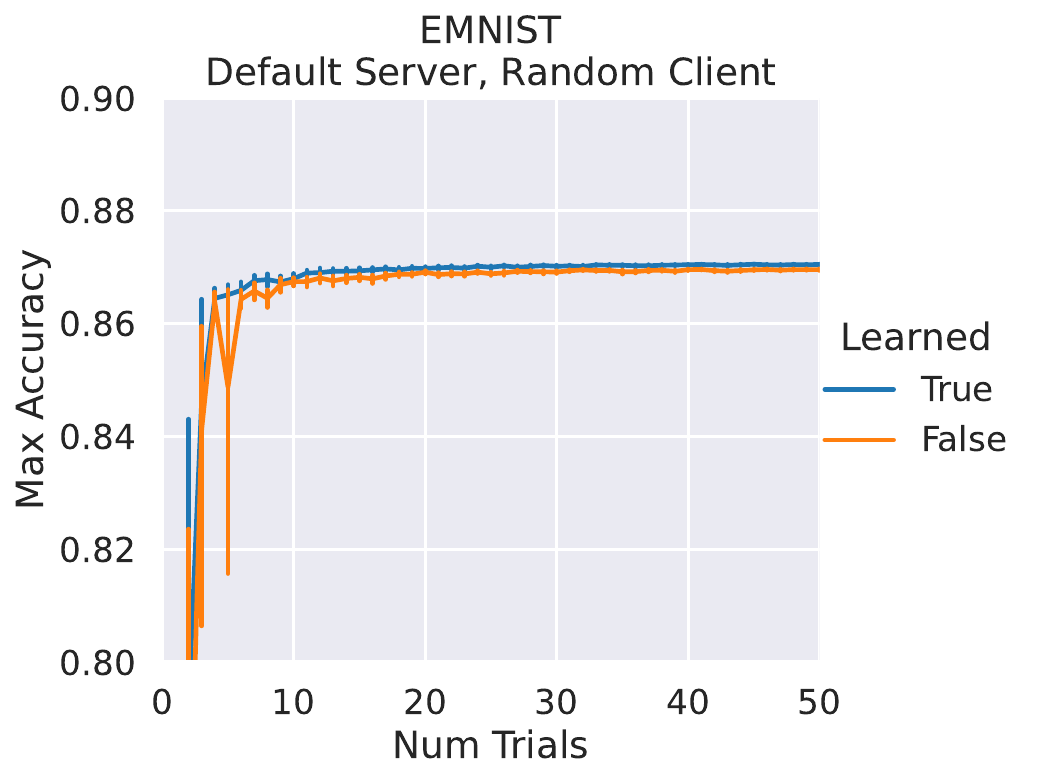}
  \label{fig:emnist}
\end{subfigure}%
\begin{subfigure}{.5\textwidth}
  \centering
    \includegraphics[width=.9\linewidth]{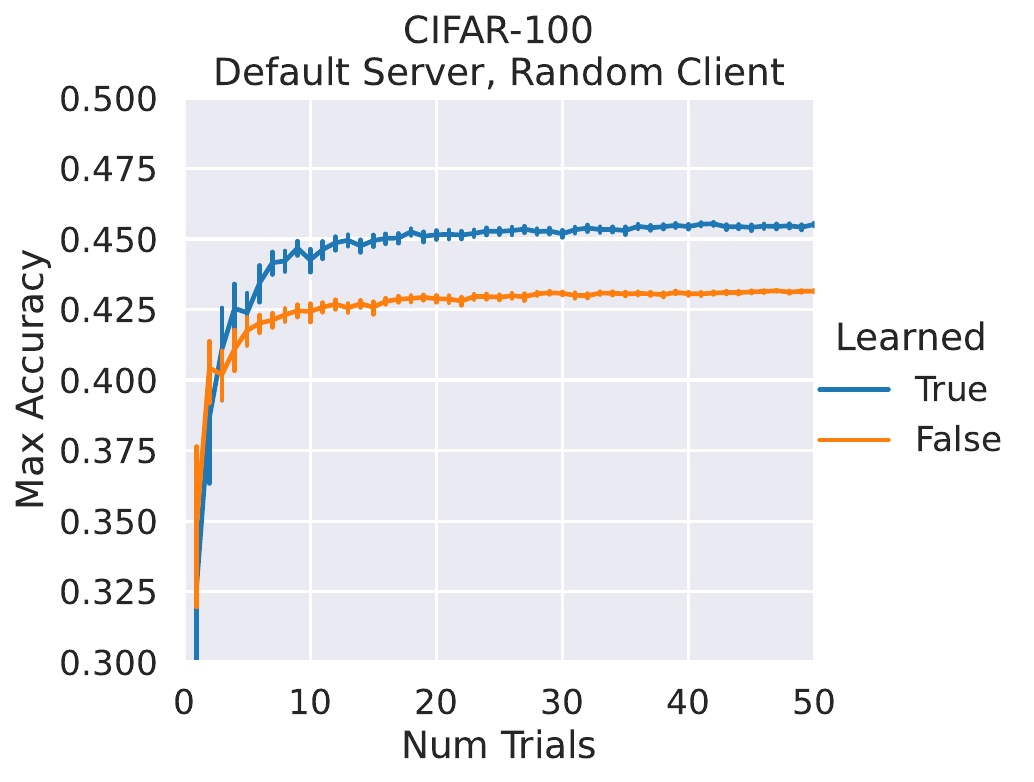}
  \label{fig:cifar}
\end{subfigure}
\centering
\begin{subfigure}{.5\textwidth}
  \centering
    \includegraphics[width=.9\linewidth]{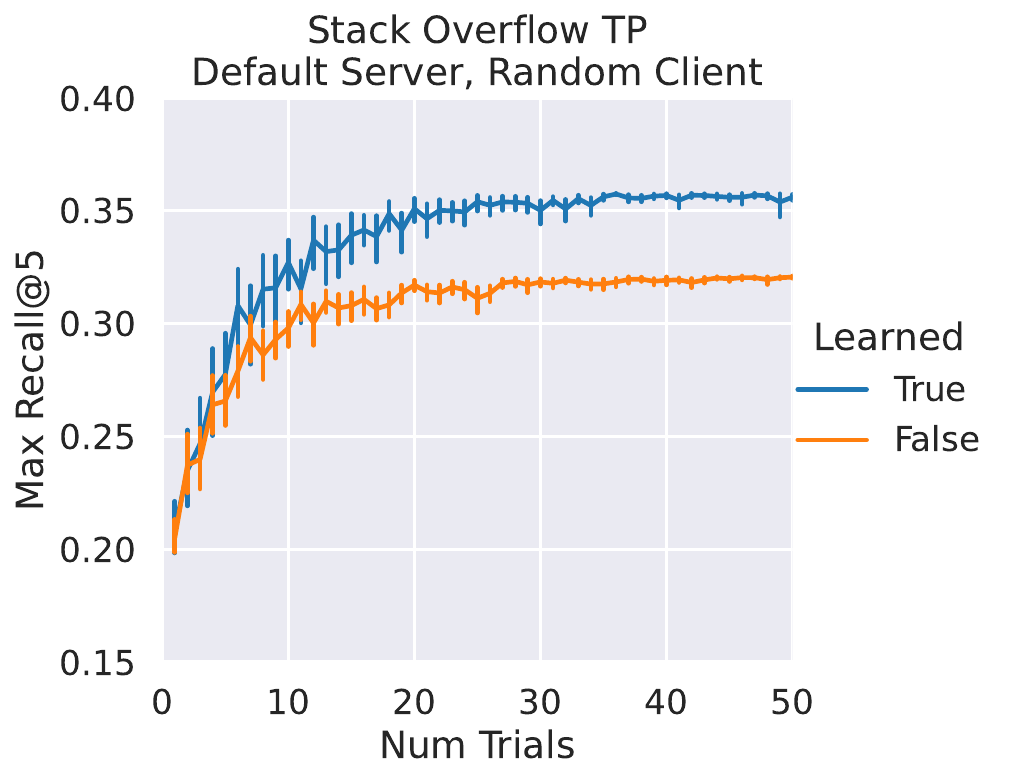}
  \label{fig:so_tag}
\end{subfigure}
\caption{Estimated max accuracy from multiple trials of default server / random client initialization, as described above. 50 different seeds were used to determine the initializations, so that fixed and learned server optimizer hyperparameters share the same set of client learning rates and the same client sampling order. The plots were generated by computing a bootstrap estimate of the maximum test accuracy across $n \leq 50$ trials.}
\label{fig:server_lr_default_sampled_plots}
\end{figure}

As shown in~\cref{tab:random_server_default_client}, allowing automatic tuning of learning rates and momentum parameters from their default values \emph{always} increases the maximum accuracy achieved in this sweep over client learning rates; again, entirely in the absence of tuning the optimizer used to compute the hyperparameter updates.

Another important aspect of learned algorithms is their ability to reduce total compute time by requiring fewer random trials. To estimate the effects of this, we use a bootstrap analysis: For each number of trials $n \leq 50$, we pick $n$ out of the 50 trials with replacement, and compute the maximum test accuracy across the $n$ trials. We do this repeatedly to produce mean and variance estimates of the maximum test accuracy attained by running $n \leq 50$ trials. We plot the results for each task in~\cref{fig:server_lr_default_sampled_plots}. We see that when running the same experimental setup with fewer samples, we can attain computational savings by doing gradient-based adjustment of the server optimizer parameters. Generally, one may expect to replicate the max accuracy from the unlearned settings with one-tenth the trials.

\begin{figure}
\begin{subfigure}{.5\textwidth}
  \centering
  \includegraphics[width=.9\linewidth]{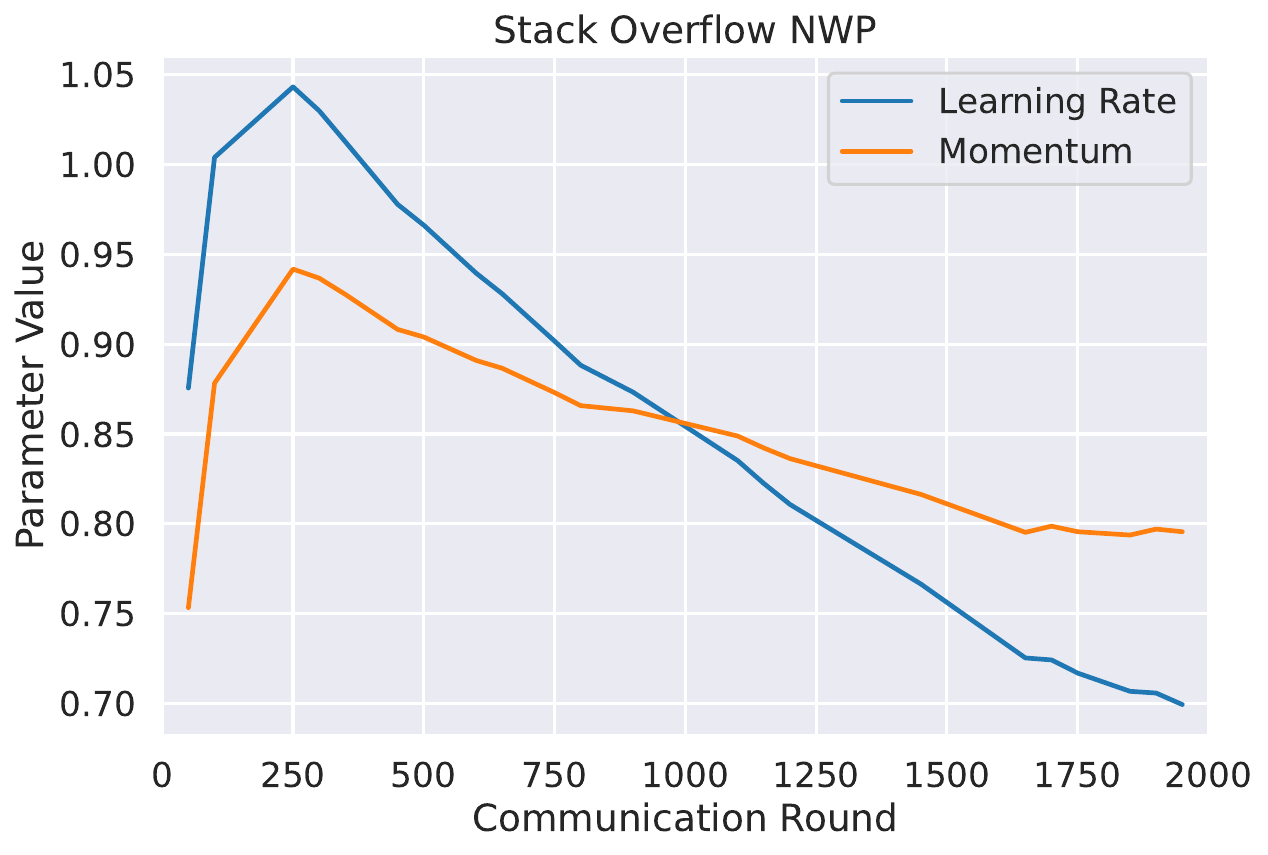}
  \label{fig:so_nwp_traj}
\end{subfigure}
\begin{subfigure}{.5\textwidth}
  \centering
  \includegraphics[width=.9\linewidth]{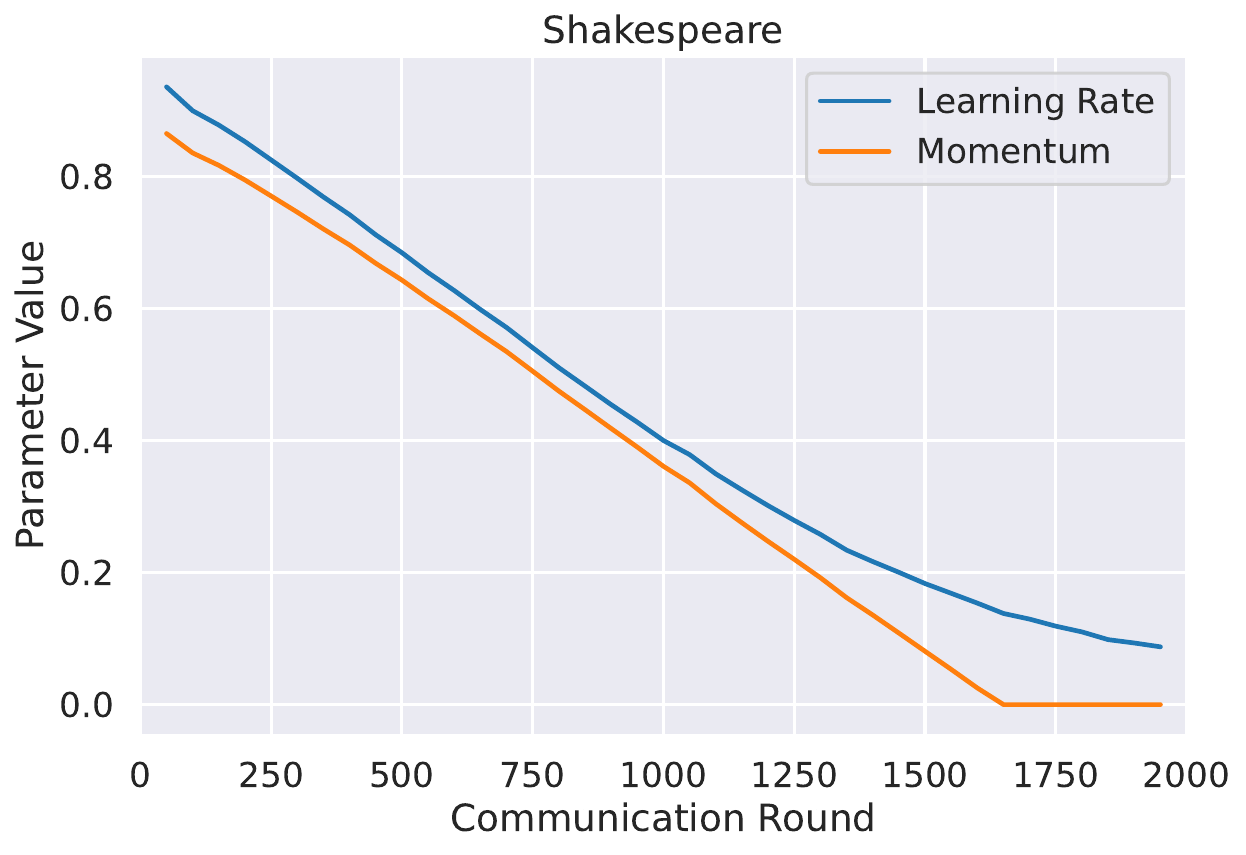}
  \label{fig:shakespeare_traj}
\end{subfigure}%
\caption{Example learned server learning rate and momentum trajectories for the Stack Overflow NWP and Shakespeare tasks.}
\label{fig:server_lr_trajectories}
\end{figure}

The trajectories of the learned learning rate and momentum parameters also yield insight into the dynamics of FL algorithms. In~\cref{fig:server_lr_trajectories}, we plot the trajectories of server learning rate and momentum parameters for Shakespeare and SO NWP trials corresponding to the optimal setting of client learning rate over our random sweep. These trajectories reflect distinct features of the tasks under consideration: SO NWP initially \emph{decreases} from its learning rate of 1.0 and momentum of 0.9, before increasing back up to slightly above these values, then slowly decaying. This initial decrease has a profound effect on training dynamics. Our default server / default client experiments, which we initialized with the optimal values of the sweep used by~\citet{afo}, were significantly more stable in the learned-hyperparameter setting: 23/50 of the non-learned experiment runs diverged very early on in training, but \emph{none} of our learned experiments diverged. Meanwhile, the Shakespeare task proved itself difficult to tune with fixed hyperparameters; low server LR values eventually outperform large values in validation accuracy, seemingly due to an analog of over-fitting, with the appropriate hyperparameter choice therefore being highly dependent on the number of rounds for which one runs one's algorithm.~\cref{fig:server_lr_trajectories} demonstrates a smooth transition from higher to lower learning rates, which enables a single learned Shakespeare task to effectively match the performance curve of multiple fixed hyperparameter specifications.

\section{Learned Client Weighting}\label{sec:learned_weight}

The parameters learned in~\cref{sec:learned_server_opt} only affected server-side computation; \fad, however, can differentiate with respect to parameters anywhere in an FL system. We now discuss how to use \fad to learn the client weights $\rho_i$ in \cref{alg:fedopt}. While these weights are fixed in \cref{alg:fedopt_hparam}, we apply federated hypergradient descent by making them a function of some hyperparameter. In order to subsume both example- and uniform-weighting, we let $\rho_i = n_i^q$ where $q$ is a smooth function of a learnable hyperparameter $\gamma$ (see \cref{app:client_weight_experimental_details} for details). Obviously, $q = 0$ recovers uniform weighting and $q = 1$ recovers example weighting. We then apply \cref{alg:fedopt_hparam} to learn $\gamma$ in tandem with the model.

We broadcast $\gamma$ to the clients at each round, who use it to compute their local weight in \cref{alg:fedopt}. We then perform the same procedure as \cref{alg:fedopt_hparam}, where we use mixed-mode \fad to compute $\partial L/\partial \gamma$, where $L$ is the empirical risk function. The corresponding federated computational graph is similar to \cref{fig:server_hparam_serial}, except that the parameter $\gamma$ governing the weight exponent is broadcast to clients. The remainder of the graph is identical. We give a full picture in \cref{fig:client_weight_serial}. This computation can also be parallelized, as in \cref{fig:server_hparam_parallel}. 

\subsection{Experimental Setup}

Now that we can apply \fad, we use \cref{alg:fedopt_hparam} to learn the model via \fedopt in tandem with the client weight parameter $\gamma$ via hypergradient descent. We apply \cref{alg:fedopt_hparam} to the same benchmark tasks in \cref{sec:learned_server_opt}, with a few minor changes. First, we special case \fedopt to \fedavg by setting \serverupdate to SGD with learning rate of $1.0$. The client learning rate is set to the tuned defaults discussed in \cref{sec:learned_server_opt}. Last, we use Adam~\citep{adam} in our hypergradient descent step (rather than SGD), with learning rate $0.01$. All other implementation details are the same.

On all 5 tasks (CIFAR-100, EMNIST, Shakespeare, SO NWP, and SO TP) we see no statistically significant difference between learned client weighting and example-weighting. For some, like the CIFAR-100 task, this is because all clients have the same number of examples. For others, the root cause of this similarity is unclear, though a large sweep over fixed parameter values for $\gamma$ also demonstrated little difference between these tasks. We therefore conjecture that these tasks are insufficiently heterogeneous to provide an effective illustration of applying \fad to learn $q$.

In order to investigate \fad for this scenario, then, we use the synthetic logistic regression problem proposed by \citet{fedprox}, Synthetic$(\alpha, \beta)$. This task has multiple forms of heterogeneity (measured by the parameters $\alpha$ and $\beta$) in addition to having unbalanced clients. Following the implementation used in \citep{fedprox}, the distribution of number of examples across clients follows a shifted log-normal distribution. For more details on this task, see \cref{app:client_weight_experimental_details}. In short, the task provides a setting where unweighted and client-weighted means perform quite differently (as noted by \citet{fedprox}), making it a useful candidate task for learned client weighting. We set $\alpha = 1, \beta = 1$ for our experiments, as this is the ``most heterogeneous'' form of the task investigated by \citet{fedprox} and \citet{fednova}.

\subsection{Results}

\begin{figure}
\begin{subfigure}{.5\textwidth}
  \centering
  \includegraphics[width=.98\linewidth]{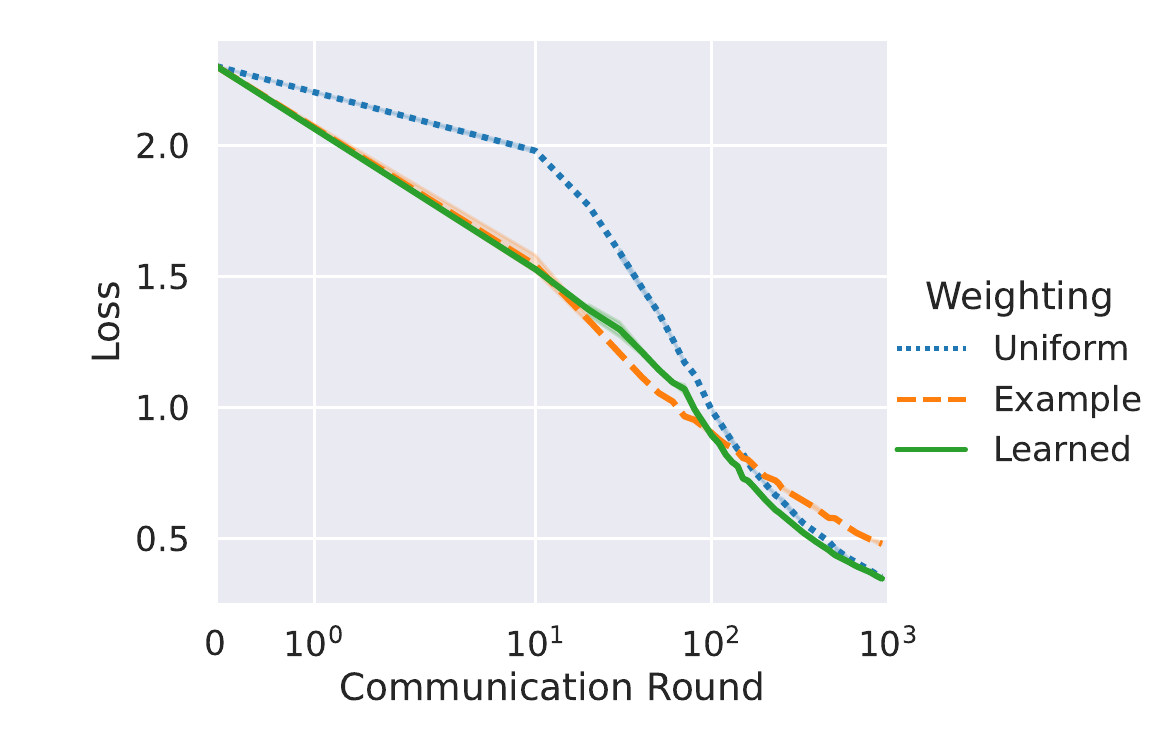}
  \label{fig:learned_client_weight_synthetic_loss}
\end{subfigure}
\begin{subfigure}{.5\textwidth}
  \centering
  \includegraphics[width=.98\linewidth]{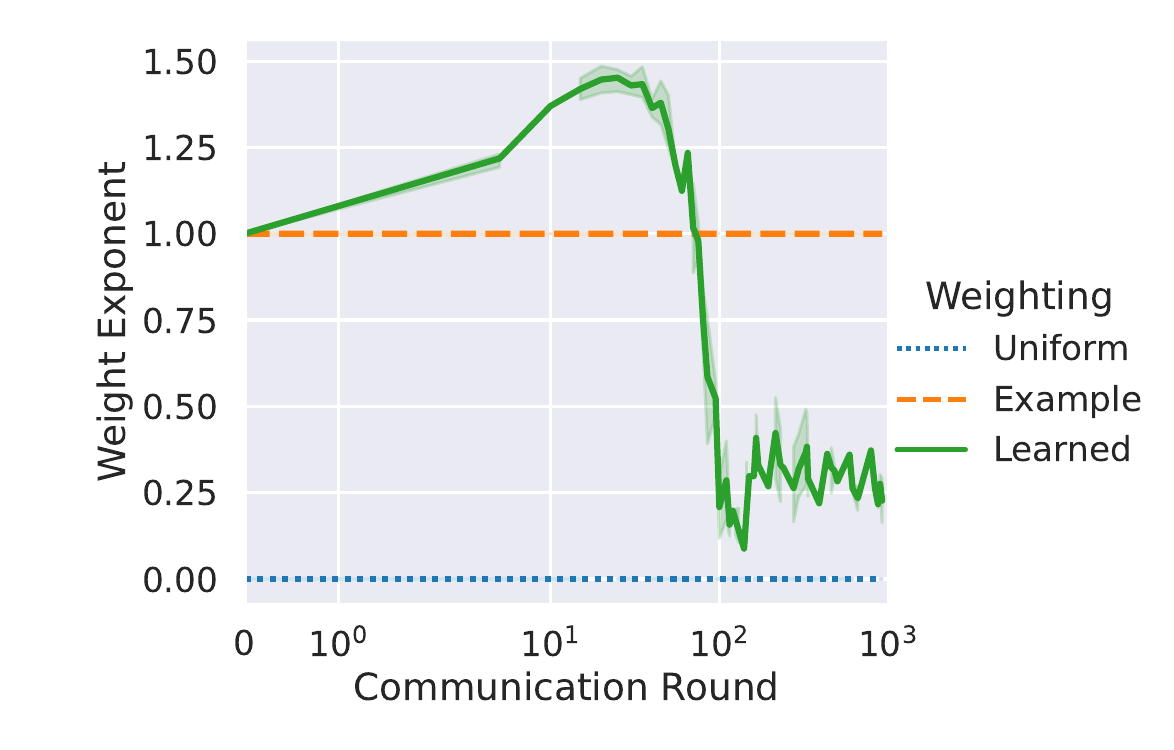}
  \label{fig:learned_client_weight_synthetic_exponent}
\end{subfigure}%
\caption{Loss and weighting exponent for the Synthetic$(1, 1)$ task, where we compare example-weighting, uniform weighting, and learned weighting.}
\label{fig:learned_client_weight_synthetic}
\end{figure}

We present our results in \cref{fig:learned_client_weight_synthetic}. Several interesting findings are worth highlighting. First, while we initialize such that the learned weighting exponent starts at $q = 1$, we see comparable (though slightly worse) results when initializing at $q = 0$ (see \cref{fig:learned_client_weight_synthetic_alt}). Second, in the fixed parameter setting, example-weighted averaging begins the training procedure by significantly outperforming uniform averaging. This phenomenon may be understood by noting that example-weighted means give higher weight to clients which took the most steps, and are therefore likely to have traveled furthest; updates therefore tend to be \emph{larger} early on in training for example-weighted \fedavg. However, the heterogeneity of this task leads to the eventual reversal of performance, with uniform weighting outperforming example weighting by the end of training.

Notably, by using \fad to learn the weight exponent, we can attain performance comparable to the \emph{lower envelope} of these two curves. In other words, \fad provides an automatic manner of interpolating between these two regimes in which different weighting is called for. This capacity for dynamic interpolation provides one potential solution to the thorny trade-offs between speed and accuracy in FL, studied in depth in~\citep{charles_and_konecny}.

\section{Conclusion}\label{sec:conclusion}

In this work, we describe a programmatic framework, federated automatic differentiation, for computing derivatives of a broad class of federated computations that are compatible with privacy-preserving technologies. Moreover, we showed that our framework yields a federated computation in the same class, so that we can be assured that derivatives can always be computed with formal privacy mechanisms as well. We presented multiple modes for \fad and discussed their implications for target systems. We applied \fad to federated hypergradient descent problems, and showed that it can improve accuracy and make finding good hyperparameters much less expensive.

This work is intended to be a starting point for the study of applying AD to FL. Some important open problems include developing software libraries capable of implementing \fad\footnote{Since the initial version of this work, we have developed a high-performance implementation of \fad in JAX; see~\citep{rush2024fax}.}, developing improved FL algorithms through the use of \fad, and thoroughly studying the behavior (theoretical and empirical) of ``dynamic'' FL algorithms that adjust their hyperparameters and operation in tandem with training.

\acks{We thank Brendan McMahan for comments on the manuscript; Roy Frostig for discussions on JAX's AD implementation; Wenjun Hu for thoughtful comments on distributed systems; Jascha Sohl-Dickstein for a dicussion of learned optimizers and evolution-strategies-based gradient computation; Matthew Streeter for early discussions on learned optimizers; Blaise Aguera y Arcas and Krzysztof Ostrowski for early support in conceptualizing FL as functional; and Jared Lichtarge and Shankar Kumar for helpful discussions of applications of \fad.
}

\appendix

\section{Learned Server Optimization - Details}\label{app:server_opt_experimental_details}

This section contains various details of the empirical evaluation in \cref{sec:learned_server_opt}. In particular, we cover the implementation details, hyperparameter initialization schemes, and methods for computing gradients that are necessary to recreate our empirical results.

\subsection{Hyperparameter Initialization}
As discussed in \cref{sec:learned_server_opt}, we initialized server hyperparameters (learning rate and momentum) and client hyperparameters (learning rate) either randomly, or with default settings. For random initialization, learning rates were chosen via a log-uniform distribution from the range $(10^{-3}, 10)$. Momentum values were chosen uniformly from the range $(0, 1)$. For default server optimizer settings, we used a learning rate of 1.0 and a momentum value of 0.9. For default client learning rate settings, we used the optimal values reported from the sweep in~\citep{afo}. All random-setting experiments were repeated 50 times.

\subsection{Implementation}
The implementation which backed our learned server optimization experiments was parameterized by function and Jacobian computations, relying on hand-implementations of a few limited federated AD components to compose these together and compute derivatives of model losses with respect to server optimizer hyperparameters. We instantiated these parameterizations with hand-differentiated pairs of functions representing server optimizer updates implemented in TensorFlow. The federated portions of our programs could be implemented in terms of existing symbols in TensorFlow-Federated, particularly those defining functions for computing \fedavg and \fedsgd. These computations were invoked in parallel, as represented in~\cref{fig:server_hparam_parallel}.

A subtle point of distinction may be made with respect to what precise relationship exists between the clients which feed the \fedavg computation and those which feed the \fedsgd procedure. At least three options come to mind immediately: using identical clients; using clients sampled from the same set (`population'), though using two different samples for each invocation of the two computations; finally, using clients sampled from two entirely different populations.

We consider the final two to be more compatible with current cross-device FL systems, where sampling a fresh batch of clients and asking them for the relatively lightweight computation of a single gradient is significantly cheaper than either requiring persistent client-server connections or increasing the workloads and payloads of individual clients. We generally expect any differences between leveraging a fresh sample from the same population and a sample from a completely different population to be artifacts of the task under consideration; for this reason, we focused on sampling clients for both purposes from a single population. Such `gradients from train clients' experiments populate all the figures used here. On some tasks, however, notably Shakespeare, leveraging gradients from validation clients demonstrated a remarkable ability to avoid the problem of over-fitting to the \emph{training population} which seems to be endemic to federated Shakespeare training.

\subsection{Additional Experimental Results}

Here we provide results on the other experiments run in support of \cref{sec:learned_server_opt}, where we only presented the results for random client / `default' server hyperparameter initializations.

\begin{figure}
\begin{subfigure}{.5\textwidth}
  \centering
  \includegraphics[width=.9\linewidth]{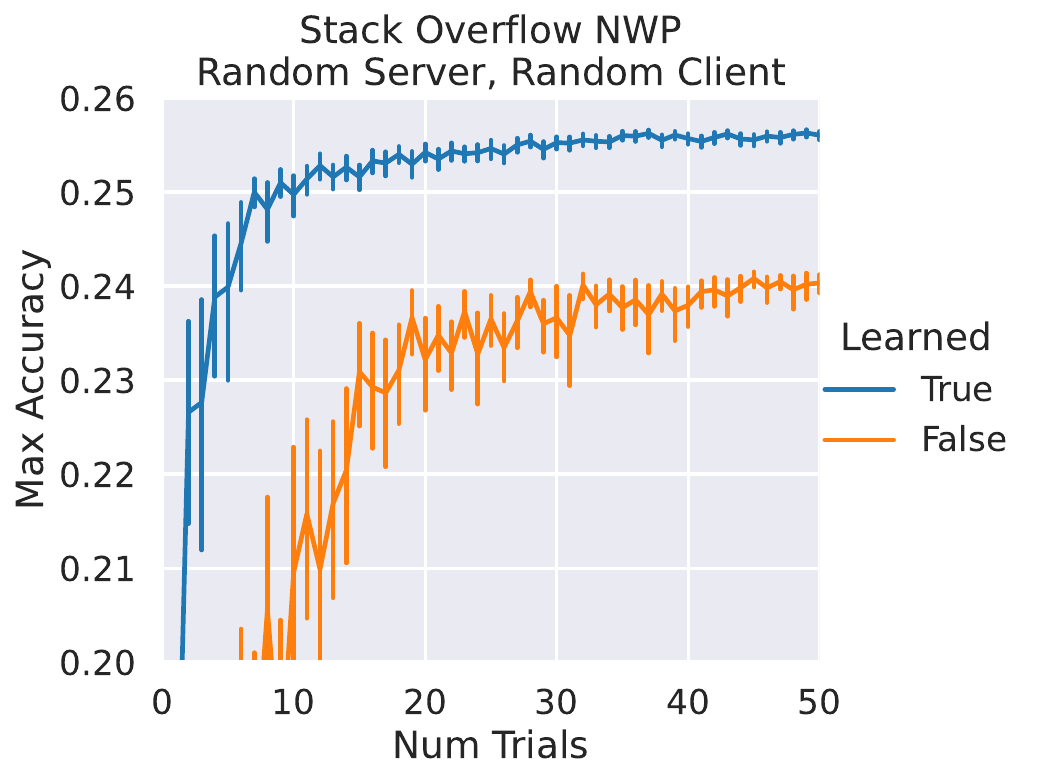}
  \label{fig:so_nwp_random_server_random_client}
\end{subfigure}%
\begin{subfigure}{.5\textwidth}
  \centering
  \includegraphics[width=.9\linewidth]{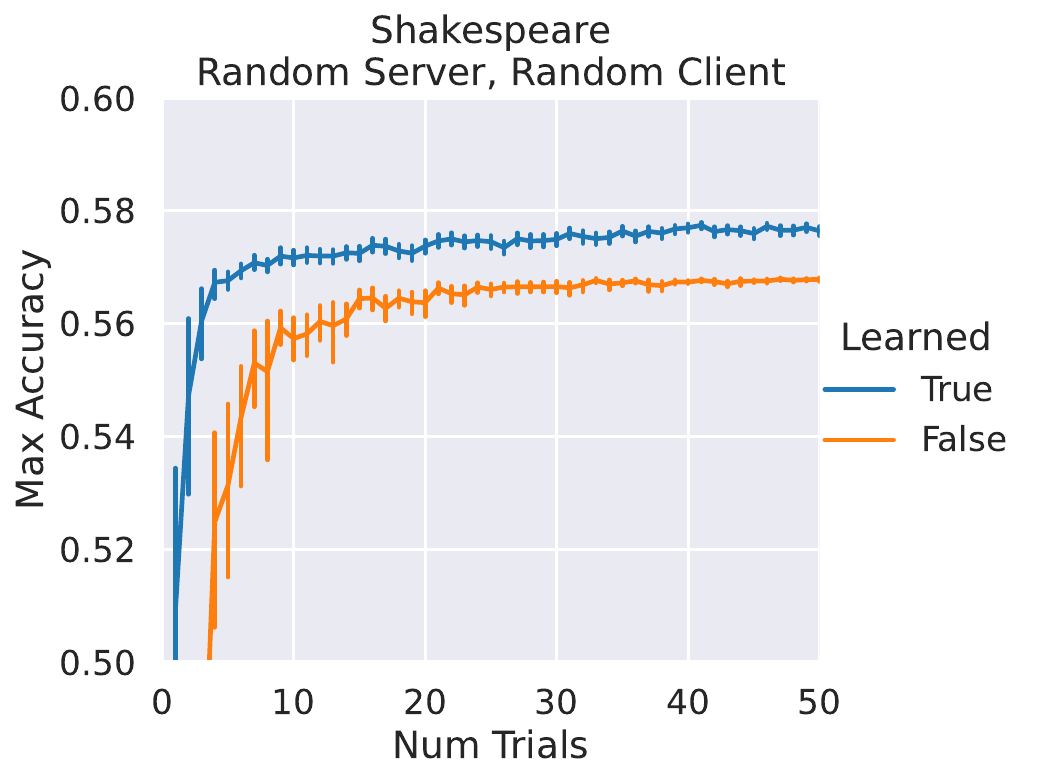}
  \label{fig:shakespeare_random_server_random_client}
\end{subfigure}
\begin{subfigure}{.5\textwidth}
  \centering
    \includegraphics[width=.9\linewidth]{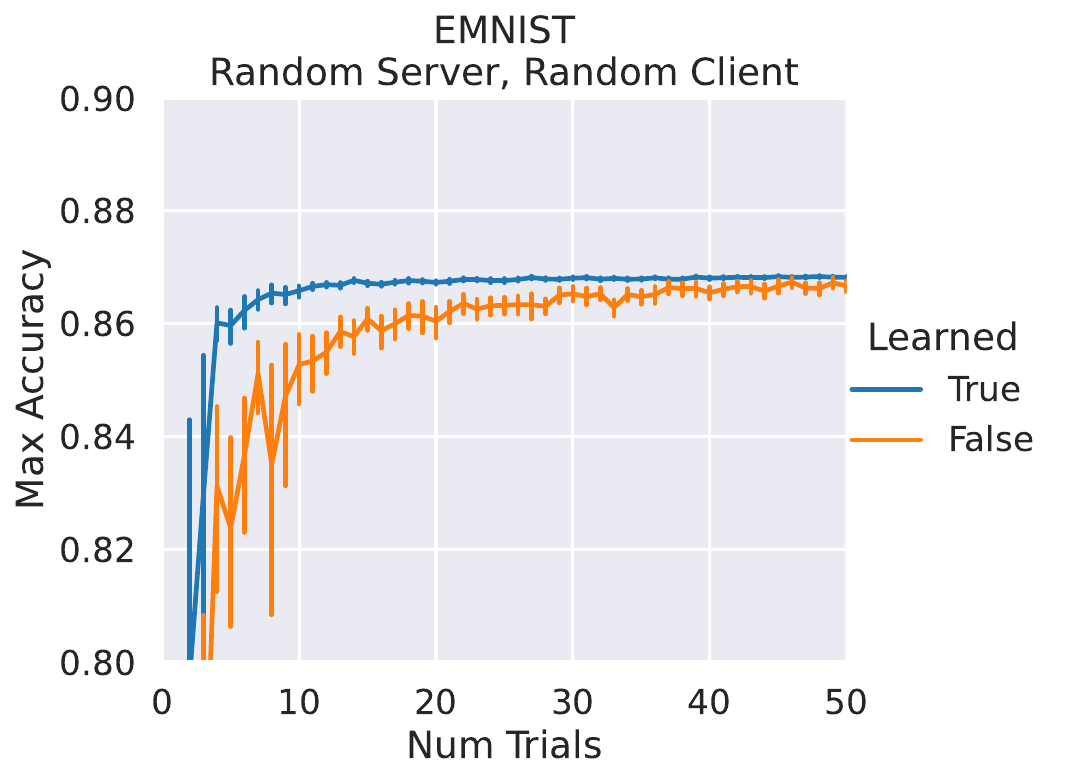}
  \label{fig:emnist_random_server_random_client}
\end{subfigure}%
\begin{subfigure}{.5\textwidth}
  \centering
    \includegraphics[width=.9\linewidth]{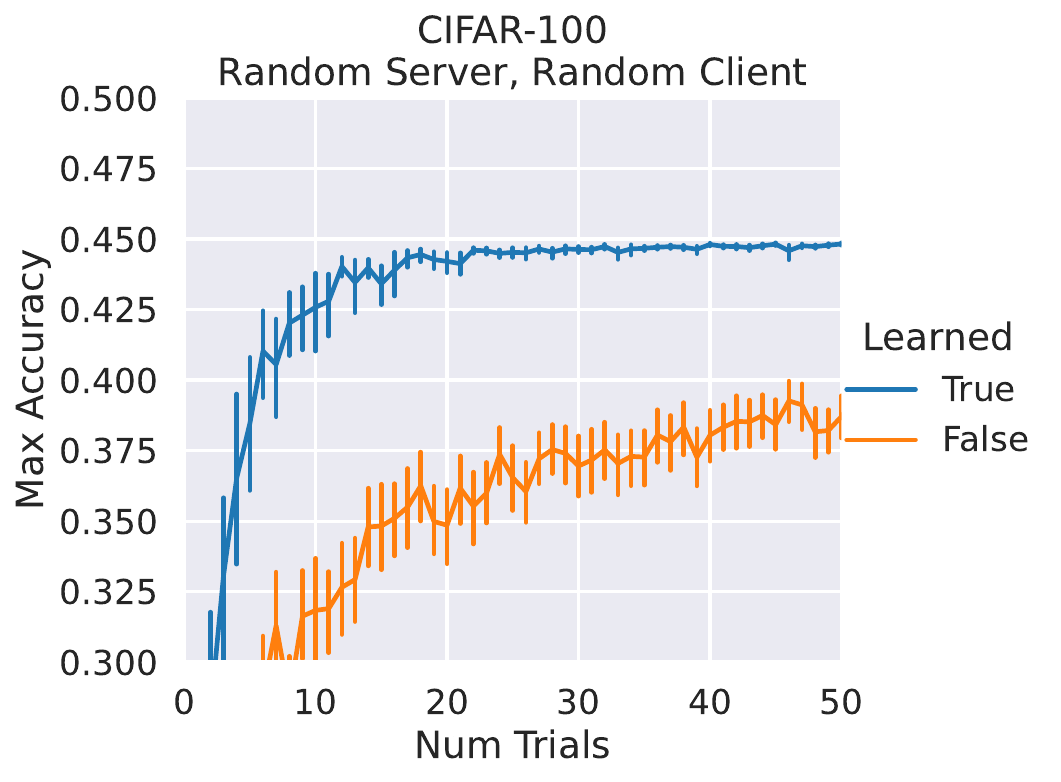}
  \label{fig:cifar_random_server_random_client}
\end{subfigure}
\centering
\begin{subfigure}{.5\textwidth}
  \centering
    \includegraphics[width=.9\linewidth]{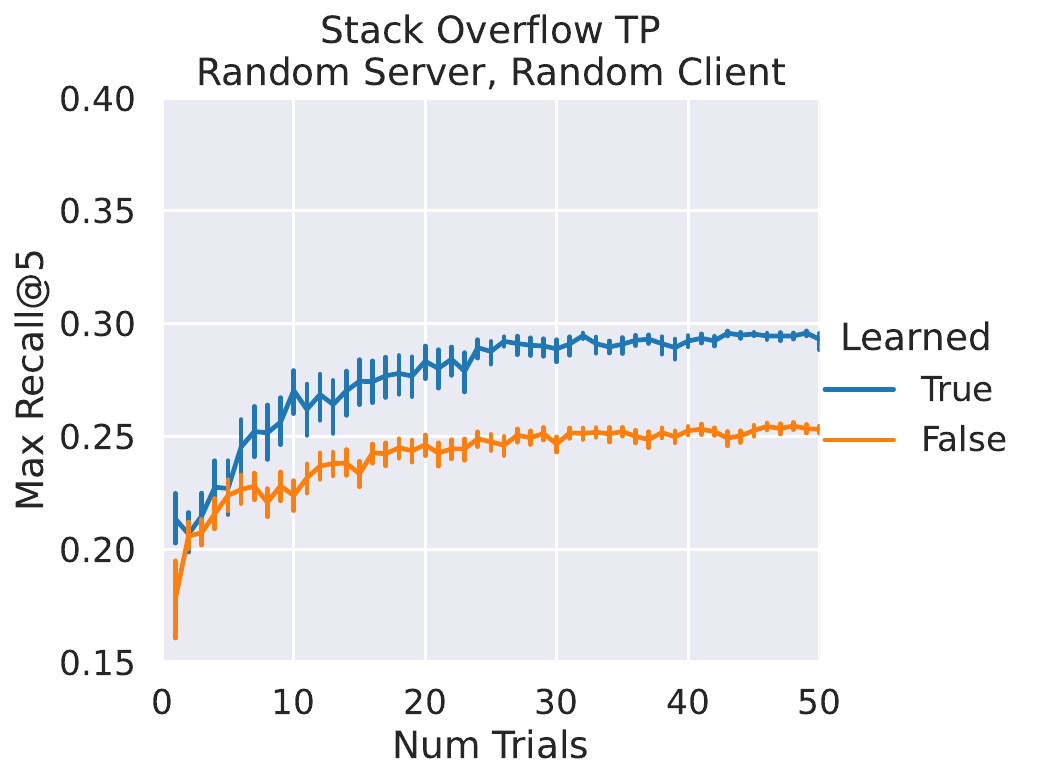}
  \label{fig:so_tag_random_server_random_client}
\end{subfigure}
\caption{Estimated max accuracy from multiple trials of random server / random client initialization, as described in \cref{sec:learned_server_opt}. 50 different seeds were used to determine the initializations, so that fixed and learned server optimizer hyperparameters share the same initial parameters and client sampling order. The plots were generated by computing a bootstrap estimate of the maximum test accuracy across $n \leq 50$ trials.}
\label{fig:random_server_random_client_sampled}
\end{figure}

\begin{figure}
\begin{subfigure}{.5\textwidth}
  \centering
  \includegraphics[width=.9\linewidth]{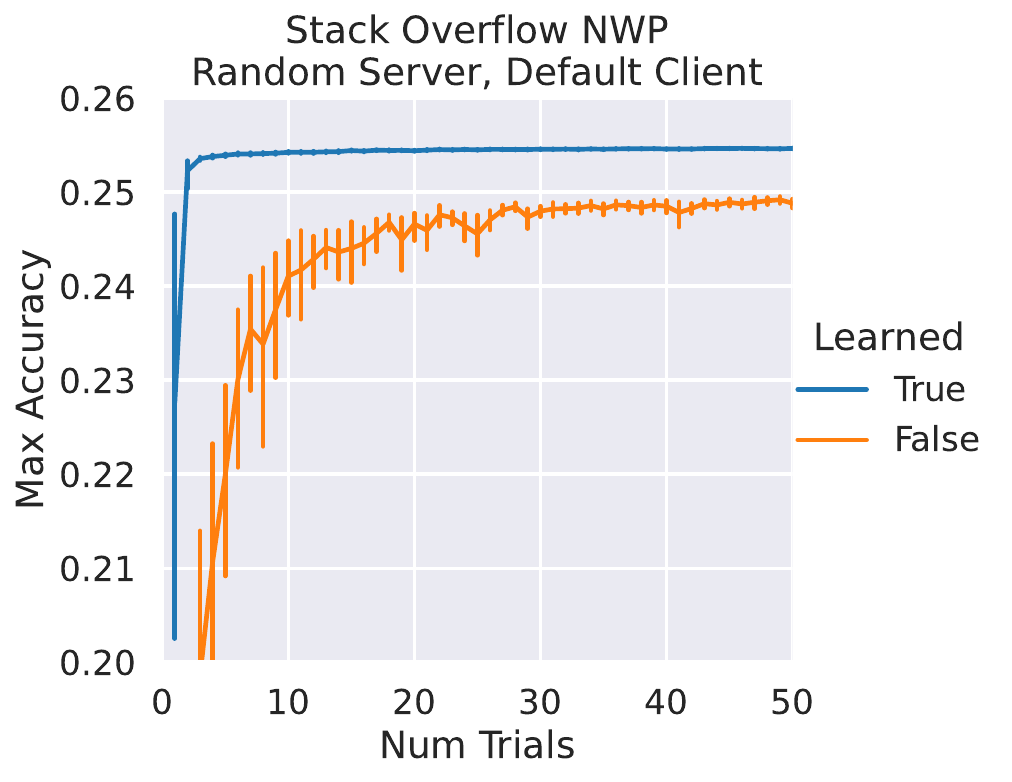}
  \label{fig:so_nwp_random_server_default_client}
\end{subfigure}%
\begin{subfigure}{.5\textwidth}
  \centering
  \includegraphics[width=.9\linewidth]{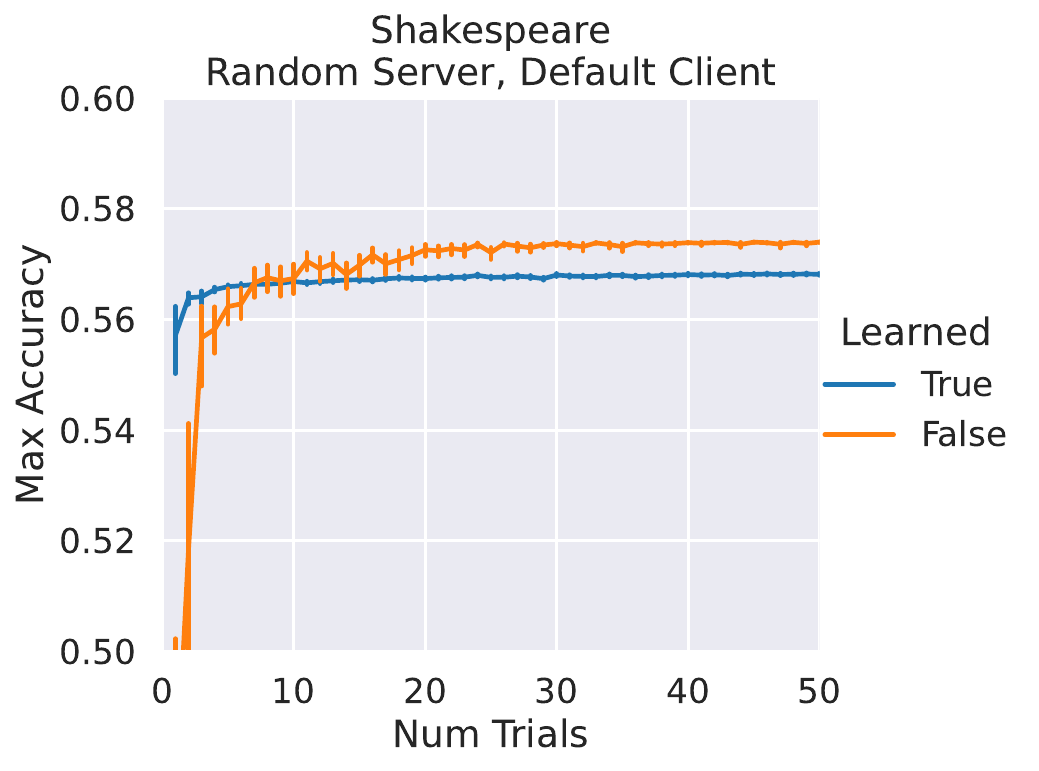}
  \label{fig:shakespeare_random_server_default_client}
\end{subfigure}
\begin{subfigure}{.5\textwidth}
  \centering
    \includegraphics[width=.9\linewidth]{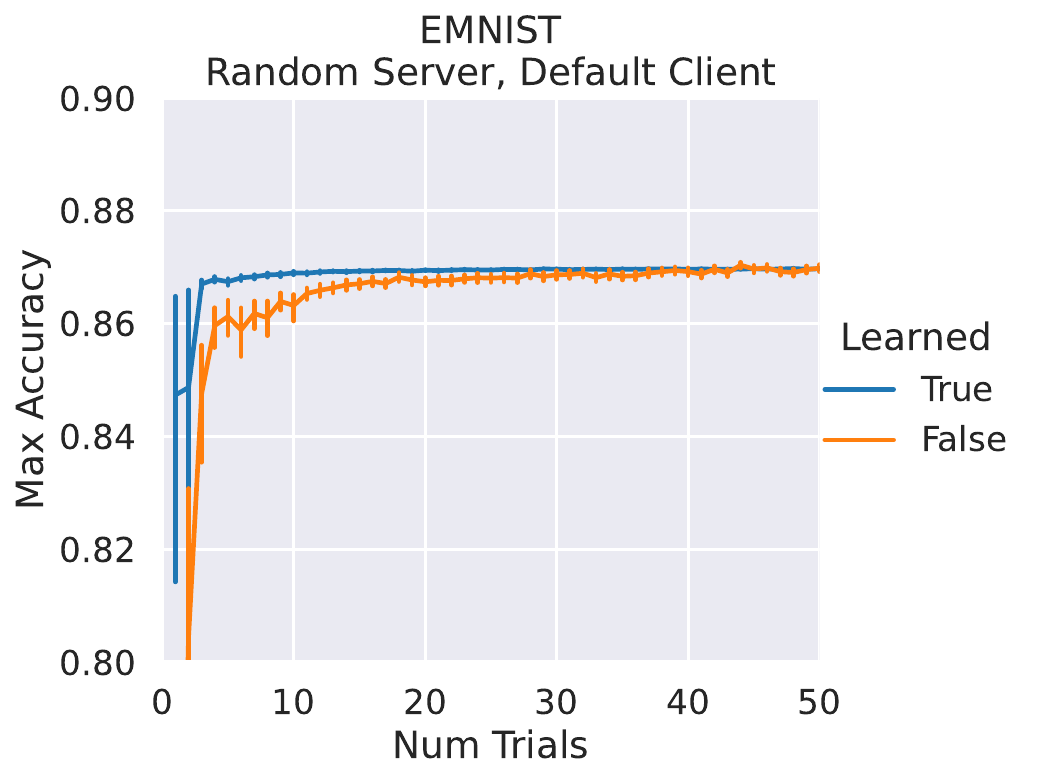}
  \label{fig:emnist_random_server_default_client}
\end{subfigure}%
\begin{subfigure}{.5\textwidth}
  \centering
    \includegraphics[width=.9\linewidth]{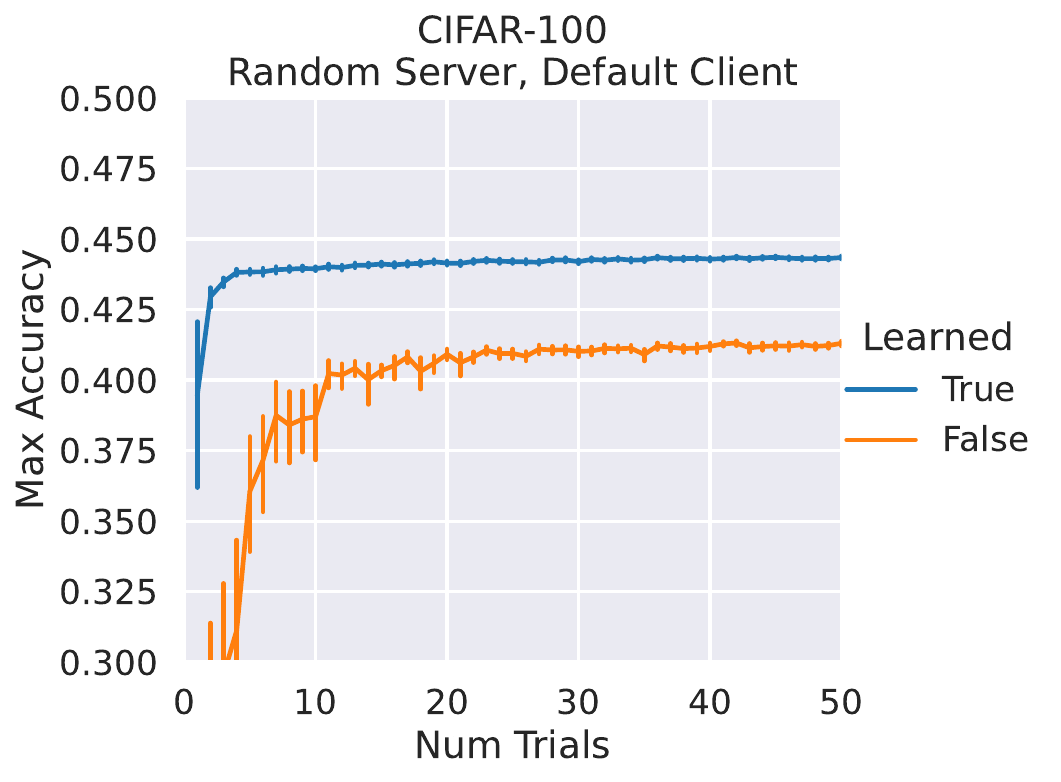}
  \label{fig:cifar_random_server_default_client}
\end{subfigure}
\centering
\begin{subfigure}{.5\textwidth}
  \centering
    \includegraphics[width=.9\linewidth]{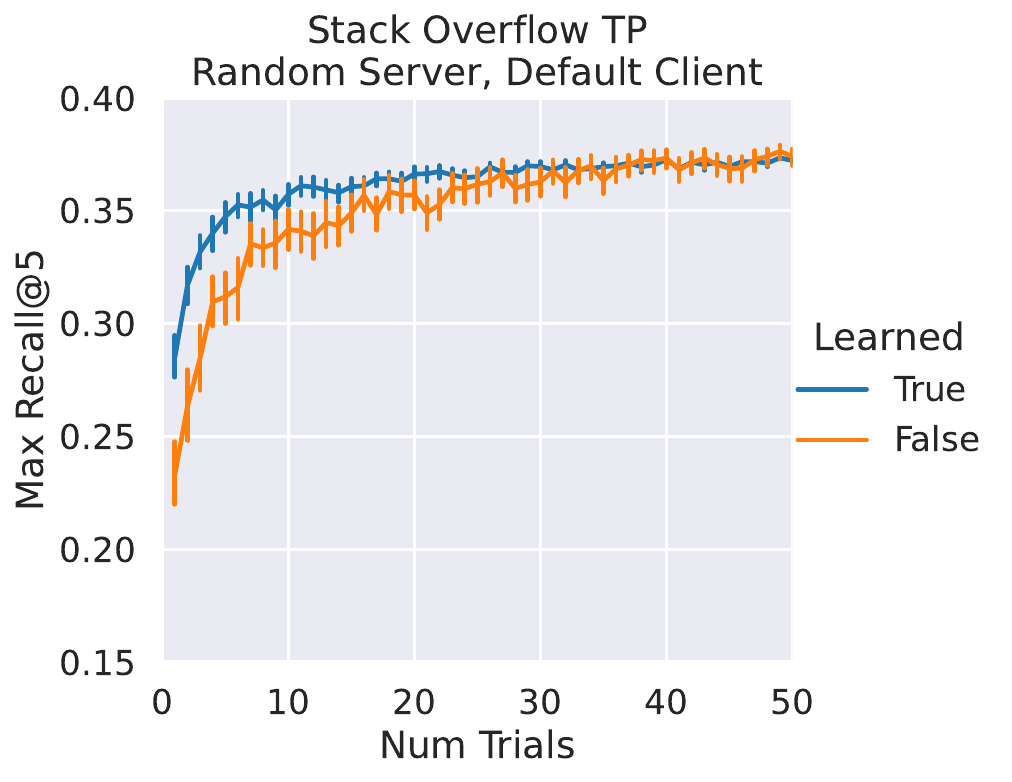}
  \label{fig:so_tag_random_server_default_client}
\end{subfigure}
\caption{Estimated max accuracy from multiple trials of random server / default client initialization, as described in \cref{sec:learned_server_opt}.}
\label{fig:random_server_default_client_sampled}
\end{figure}

\begin{figure}
\begin{subfigure}{.5\textwidth}
  \centering
  \includegraphics[width=.9\linewidth]{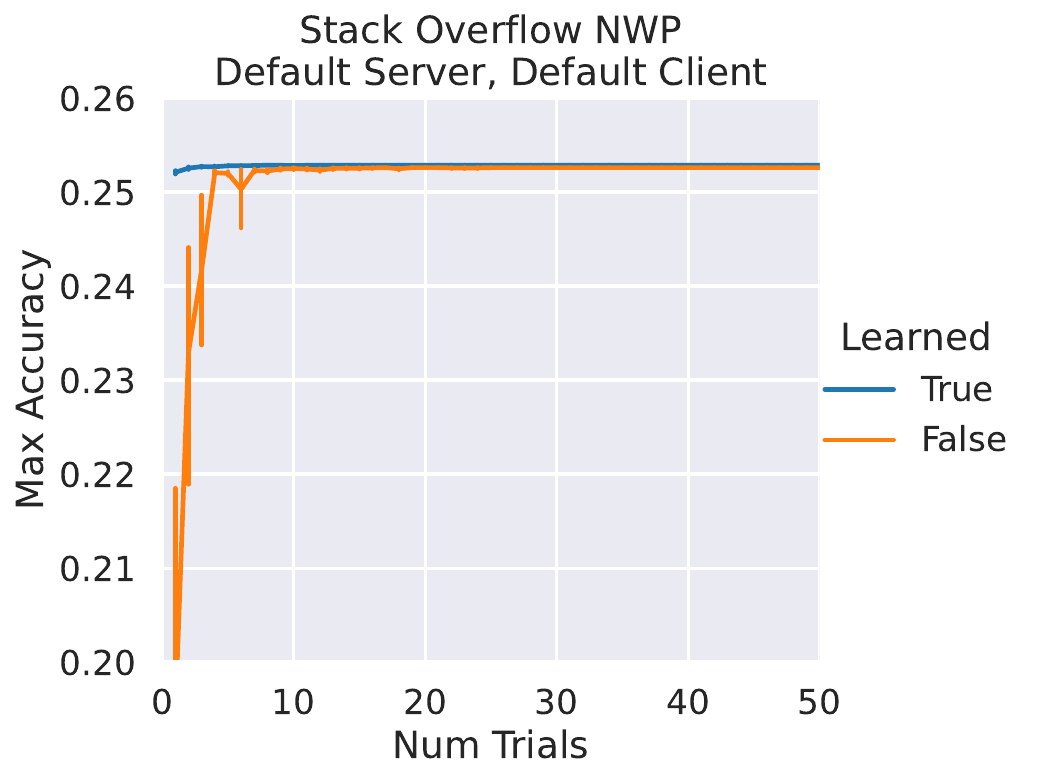}
  \label{fig:so_nwp_default_server_default_client}
\end{subfigure}%
\begin{subfigure}{.5\textwidth}
  \centering
  \includegraphics[width=.9\linewidth]{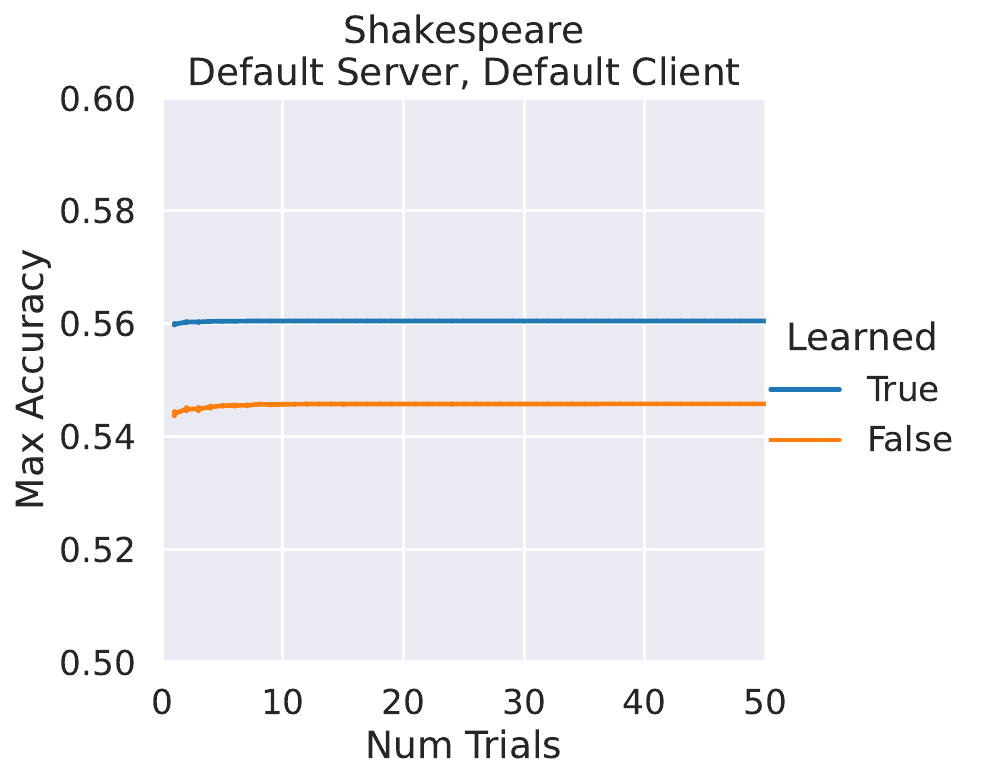}
  \label{fig:shakespeare_default_server_default_client}
\end{subfigure}
\begin{subfigure}{.5\textwidth}
  \centering
    \includegraphics[width=.9\linewidth]{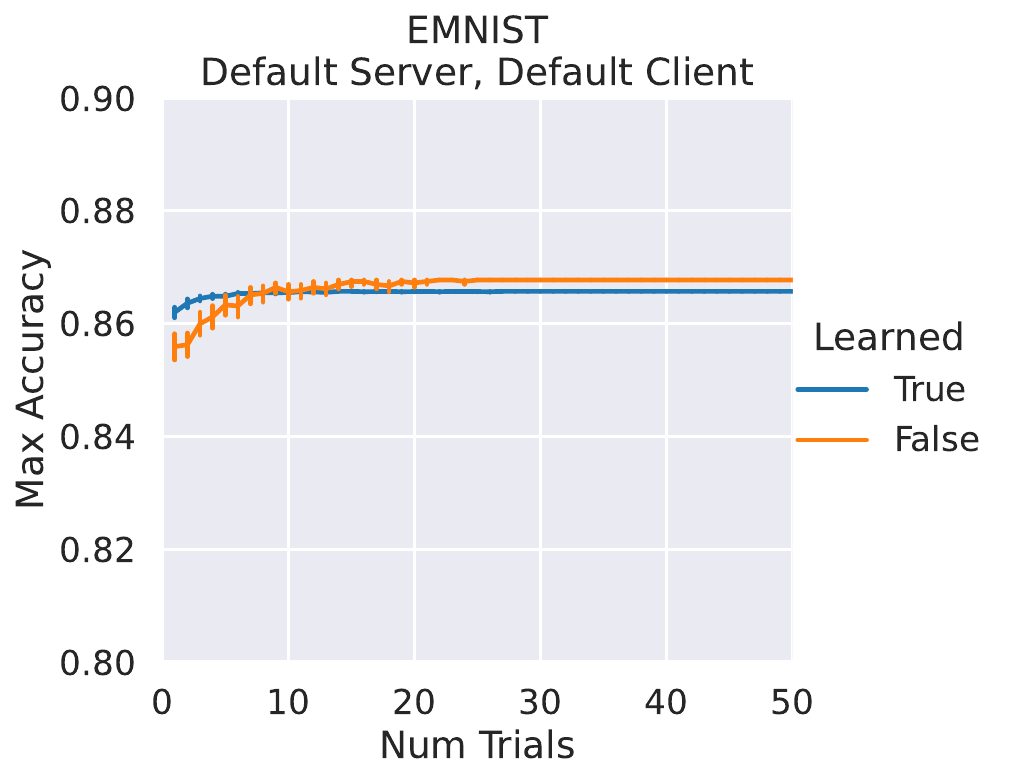}
  \label{fig:emnist_default_server_default_client}
\end{subfigure}%
\begin{subfigure}{.5\textwidth}
  \centering
    \includegraphics[width=.9\linewidth]{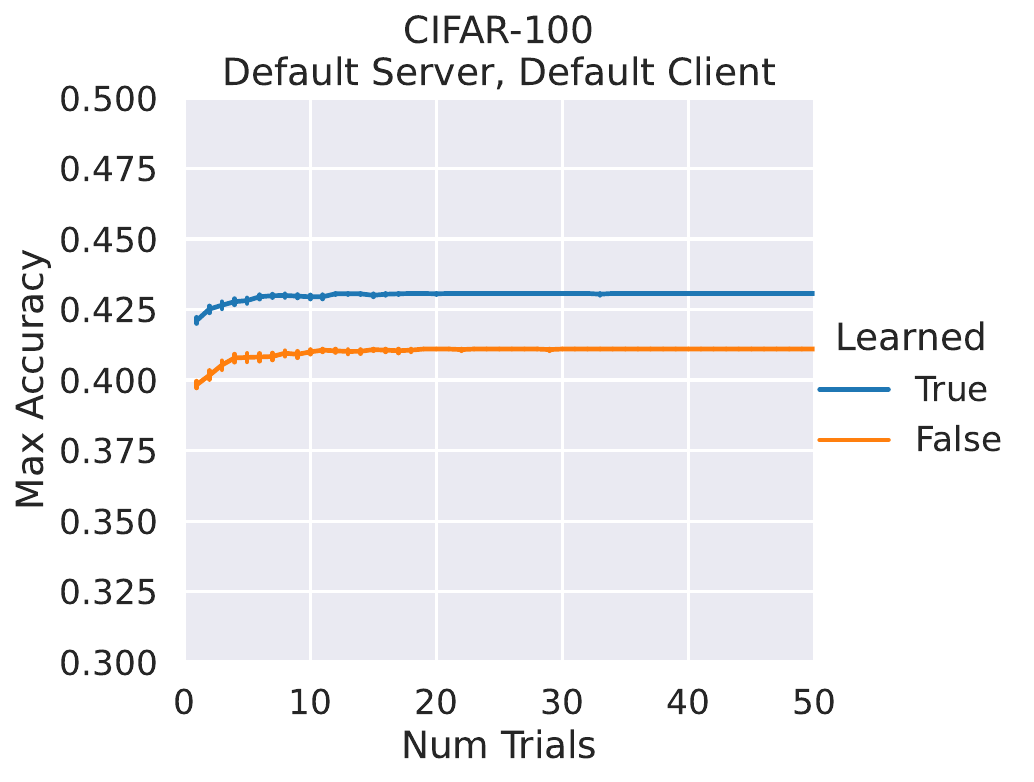}
  \label{fig:cifar_default_server_default_client}
\end{subfigure}
\centering
\begin{subfigure}{.5\textwidth}
  \centering
    \includegraphics[width=.9\linewidth]{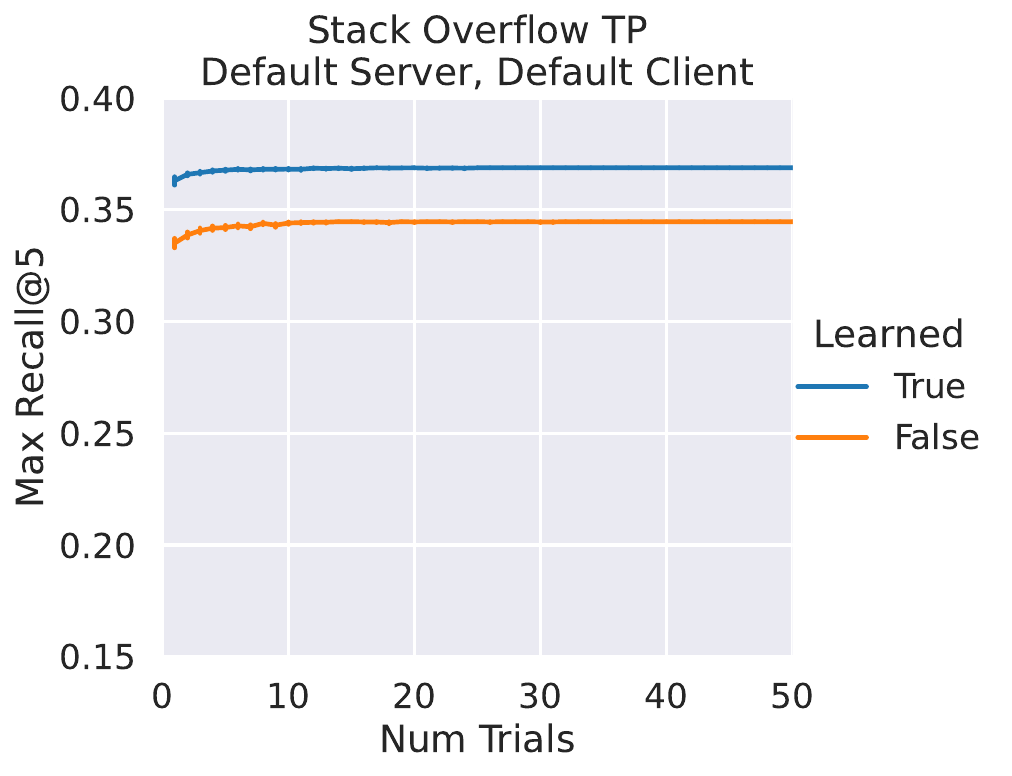}
  \label{fig:so_tag_default_server_default_client}
\end{subfigure}
\caption{Estimated max accuracy from multiple trials of default server / default client initialization, as described in \cref{sec:learned_server_opt}.}
\label{fig:default_server_default_client_sampled}
\end{figure}

Below we present the results for the other three combinations (random server/random client, random server/default client, and default server/default client). As in \cref{sec:learned_server_opt}, we run 50 trials for each (across five tasks), and use a bootstrap analysis to estimate the maximum test accuracy when running $n \leq 50$ trials. The results are given in Figures \ref{fig:random_server_random_client_sampled}, \ref{fig:random_server_default_client_sampled}, and \ref{fig:default_server_default_client_sampled}. In nearly all cases, we see that the learned server optimizer can attain a higher maximum accuracy in many fewer trials.

\clearpage

\section{Learned Client Weighting - Details}\label{app:client_weight_experimental_details}

In this section, we discuss details concerning the experimental investigations in \cref{sec:learned_weight}. In particular, we provide a complete description of the computational graph to which we apply \fad, details on the synthetic data set used, and additional experimental results.

\subsection{Federated Computational Graph}

\begin{figure}[t]
\caption{Federated computational graph used to compute hypergradients of the client weight exponent used in \cref{sec:learned_weight}. All server$\to$client communication is done via \fedbroadcast, while all client$\to$server is done via \fedmean. This computation produces some updated model and an estimate of the loss of that model. By applying \fad, we can compute the desired hypergradients.}
\includegraphics[width=0.98\linewidth]{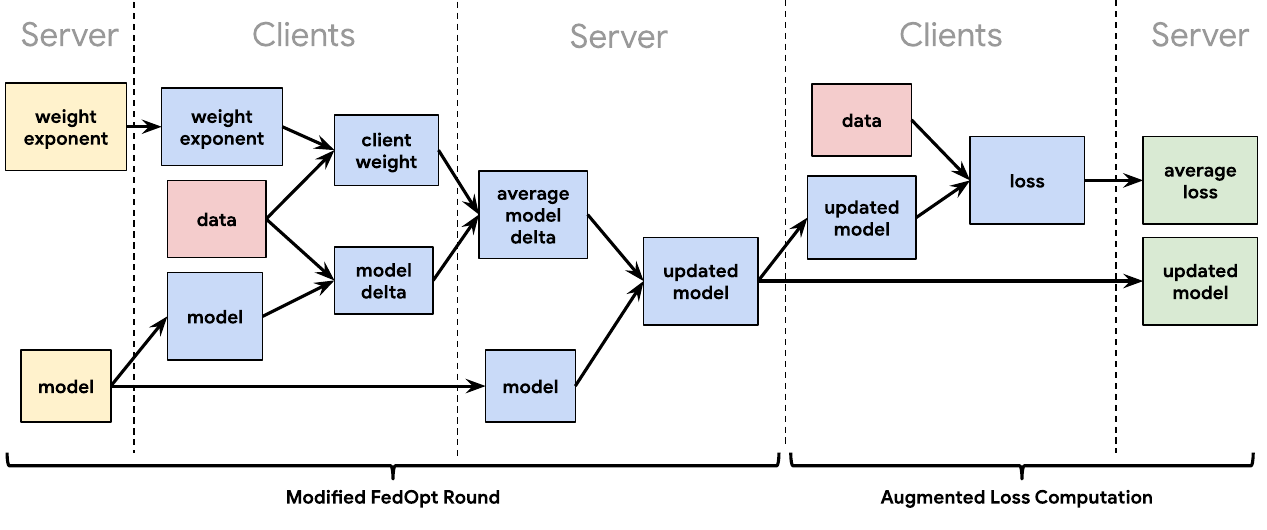}
\centering
\label{fig:client_weight_serial}
\end{figure}

Here we provide a computational graph demonstrating the parameterization of client weighting as discussed in \cref{sec:learned_weight}.The approach largely mirrors the approach used to apply \fad to server optimizer hyperparameters in \cref{fig:server_hparam_serial}. The key difference is that the client weight parameter must be broadcast to the clients at each round, since the server updates it between rounds. The resulting procedure is given in \cref{fig:client_weight_serial}. Note that this can be parallelized in an almost identical way to that of \cref{fig:server_hparam_parallel}.

\subsection{Synthetic Logistic Regression Task}

We use the $Synthetic(\alpha, \beta)$ task proposed by \citet{fedprox}. Each client $i$ has the following parameters sampled independently:
\begin{itemize}
    \item A number of example $n_i \sim \mathcal{P}$.
    \item A model parameter $u_i\sim\normal(0, \alpha)$.
    \item A mean vector $v_i \in \R^{60}$ with elements drawn from $\normal(\beta'_i, 1)$ where $\beta'_i \sim \normal(0, \beta)$.
\end{itemize}

Each client $i$ has a ``ground truth'' multi-class logistic regression model $f(x) = \argmax(\sigma(W_ix + b_i))$ where $\sigma$ is the softmax function, $W_i \in \R^{10\times 60}$, and $b_i \in \R^{10}$. The entries of $W_i$ and $b_i$ are sampled independently from $\normal(u_i, 1)$. Each client generates data $\{(x_{i, n}, y_{i, n})\}_{n=1}^{n_i}$ via $x_{i, n}\sim\normal(v_i, \Sigma)$, $y_{i, n} = f(x_{i, n})$. Here, $\Sigma$ is a diagonal covariance matrix with $\Sigma_{j, j} = j^{-1.2}$.

The remaining detail is the distribution $\mathcal{P}$ for the number of examples per client. We use the same implementation from \citet{fedprox}, where $\mathcal{P}$ is a shifted log-normal distribution. Specifically, $n_i = 50 + \lfloor\exp(m_i)\rfloor$ where $m_i \sim \normal(4, 2)$. In our empirical evaluation in \cref{sec:learned_weight}, we let there be 100 clients overall, and sample 50 of them at each round.

\subsection{Additional Experimental Results}

\begin{figure}[t]
\begin{subfigure}{.5\textwidth}
  \centering
  \includegraphics[width=.98\linewidth]{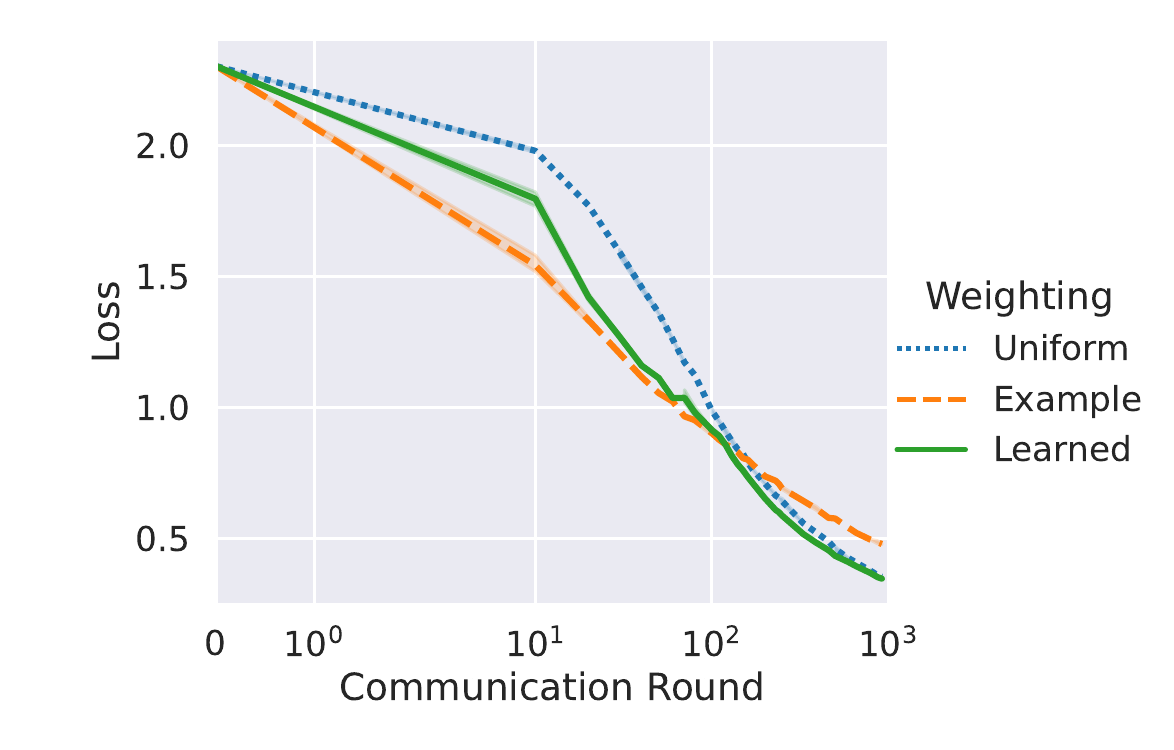}
  \label{fig:learned_client_weight_synthetic_loss_alt}
\end{subfigure}
\begin{subfigure}{.5\textwidth}
  \centering
  \includegraphics[width=.98\linewidth]{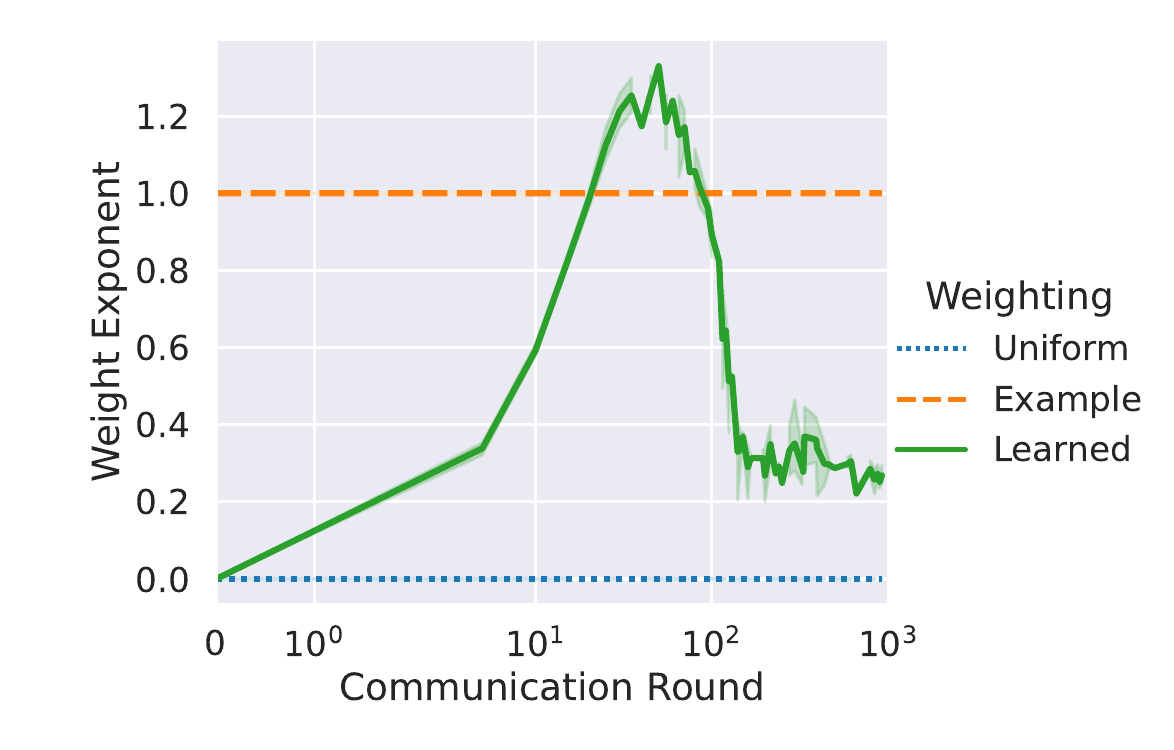}
  \label{fig:learned_client_weight_synthetic_exponent_alt}
\end{subfigure}%
\caption{Loss and weighting exponent for the Synthetic$(1, 1)$ task, where we compare example-weighting, uniform weighting, and learned weighting. While we initialize the learned weighting exponent at $q = 0$, we see comparable (though slightly better) results when initializing at $q = 1$ (see \cref{fig:learned_client_weight_synthetic}).}
\label{fig:learned_client_weight_synthetic_alt}
\end{figure}

As in \cref{sec:learned_weight}, we compare uniform weighting, example weighting, and learned weighting variants of the \fedopt algorithm (\cref{alg:fedopt}) on the synthetic logistic regression task detailed above. While \cref{fig:learned_client_weight_synthetic} compared the three in a regime where the learned weight starts off at example weighting, we compare the three when the learned weight starts off at uniform weighting in \cref{fig:learned_client_weight_synthetic_alt}.

We see that the learned weighting does better than the uniform initially (though there is some gap with example weighting) but eventually does better than both (and significantly better than example weighting, which levels off). Looking at the learned weight exponent, it quickly increases initially (in fact, to a weighting scheme where clients with more examples are weighted more heavily) and then goes back to some thing closer to uniform weighting eventually. In short, the learned weighting seems to approximate the lower envelope of the loss for uniform and example weighting.

\vskip 0.2in
\bibliography{references}

\begin{thebibliography}{71}
\providecommand{\natexlab}[1]{#1}
\providecommand{\url}[1]{\texttt{#1}}
\expandafter\ifx\csname urlstyle\endcsname\relax
  \providecommand{\doi}[1]{doi: #1}\else
  \providecommand{\doi}{doi: \begingroup \urlstyle{rm}\Url}\fi

\bibitem[Abadi et~al.(2016)Abadi, Barham, Chen, Chen, Davis, Dean, Devin, Ghemawat, Irving, Isard, et~al.]{tensorflow}
Mart{\'\i}n Abadi, Paul Barham, Jianmin Chen, Zhifeng Chen, Andy Davis, Jeffrey Dean, Matthieu Devin, Sanjay Ghemawat, Geoffrey Irving, Michael Isard, et~al.
\newblock Tensor{F}low: a system for large-scale machine learning.
\newblock In \emph{12th USENIX symposium on operating systems design and implementation (OSDI 16)}, pages 265--283, 2016.

\bibitem[Andrychowicz et~al.(2016)Andrychowicz, Denil, Colmenarejo, Hoffman, Pfau, Schaul, Shillingford, and de~Freitas]{learning2learn}
Marcin Andrychowicz, Misha Denil, Sergio~G\'{o}mez Colmenarejo, Matthew~W. Hoffman, David Pfau, Tom Schaul, Brendan Shillingford, and Nando de~Freitas.
\newblock Learning to learn by gradient descent by gradient descent.
\newblock In \emph{Proceedings of the 30th International Conference on Neural Information Processing Systems}, NIPS'16, page 3988–3996, Red Hook, NY, USA, 2016. Curran Associates Inc.
\newblock ISBN 9781510838819.

\bibitem[Authors(2019)]{stackoverflow}
The TensorFlow~Federated Authors.
\newblock Tensor{F}low {F}ederated {Stack Overflow} dataset, 2019.
\newblock URL \url{https://www.tensorflow.org/federated/api_docs/python/tff/simulation/datasets/stackoverflow/load_data}.

\bibitem[Bauer(1974)]{computational_graph}
Friedrich~L Bauer.
\newblock Computational graphs and rounding error.
\newblock \emph{SIAM Journal on Numerical Analysis}, 11\penalty0 (1):\penalty0 87--96, 1974.

\bibitem[Baydin et~al.(2018{\natexlab{a}})Baydin, Cornish, Mart{\'{\i}}nez{-}Rubio, Schmidt, and Wood]{hypergrad_descent}
Atilim~Gunes Baydin, Robert Cornish, David Mart{\'{\i}}nez{-}Rubio, Mark Schmidt, and Frank Wood.
\newblock Online learning rate adaptation with hypergradient descent.
\newblock In \emph{6th International Conference on Learning Representations, {ICLR} 2018, Vancouver, BC, Canada, April 30 - May 3, 2018, Conference Track Proceedings}. OpenReview.net, 2018{\natexlab{a}}.
\newblock URL \url{https://openreview.net/forum?id=BkrsAzWAb}.

\bibitem[Baydin et~al.(2018{\natexlab{b}})Baydin, Pearlmutter, Radul, and Siskind]{autodiff_survey}
Atilim~Gunes Baydin, Barak~A. Pearlmutter, Alexey~Andreyevich Radul, and Jeffrey~Mark Siskind.
\newblock Automatic differentiation in machine learning: a survey.
\newblock \emph{Journal of Machine Learning Research}, 18\penalty0 (153):\penalty0 1--43, 2018{\natexlab{b}}.
\newblock URL \url{http://jmlr.org/papers/v18/17-468.html}.

\bibitem[Bello et~al.(2017)Bello, Zoph, Vasudevan, and Le]{bello17}
Irwan Bello, Barret Zoph, Vijay Vasudevan, and Quoc~V. Le.
\newblock Neural optimizer search with reinforcement learning.
\newblock In Doina Precup and Yee~Whye Teh, editors, \emph{Proceedings of the 34th International Conference on Machine Learning, {ICML} 2017, Sydney, NSW, Australia, 6-11 August 2017}, volume~70 of \emph{Proceedings of Machine Learning Research}, pages 459--468. {PMLR}, 2017.
\newblock URL \url{http://proceedings.mlr.press/v70/bello17a.html}.

\bibitem[Bengio et~al.(1991)Bengio, Bengio, and Cloutier]{learning_synaptic_update}
Y~Bengio, S~Bengio, and J~Cloutier.
\newblock Learning a synaptic learning rule.
\newblock In \emph{IJCNN-91-Seattle International Joint Conference on Neural Networks}, volume~2, pages 969--vol. IEEE, 1991.

\bibitem[Bengio(2000)]{bengio_grad_hparam_opt}
Yoshua Bengio.
\newblock Gradient-based optimization of hyperparameters.
\newblock \emph{Neural Computation}, 12\penalty0 (8):\penalty0 1889--1900, 2000.
\newblock \doi{10.1162/089976600300015187}.

\bibitem[Bonawitz et~al.(2017)Bonawitz, Ivanov, Kreuter, Marcedone, McMahan, Patel, Ramage, Segal, and Seth]{secagg}
K.A. Bonawitz, Vladimir Ivanov, Ben Kreuter, Antonio Marcedone, H.~Brendan McMahan, Sarvar Patel, Daniel Ramage, Aaron Segal, and Karn Seth.
\newblock Practical secure aggregation for privacy-preserving machine learning.
\newblock In \emph{Proceedings of the 2017 ACM SIGSAC Conference on Computer and Communications Security}, CCS '17, page 1175–1191, New York, NY, USA, 2017. Association for Computing Machinery.
\newblock ISBN 9781450349468.
\newblock \doi{10.1145/3133956.3133982}.
\newblock URL \url{https://doi.org/10.1145/3133956.3133982}.

\bibitem[Bonawitz et~al.(2019)Bonawitz, Eichner, Grieskamp, Huba, Ingerman, Ivanov, Kiddon, Kone\v{c}n\'{y}, Mazzocchi, McMahan, Van~Overveldt, Petrou, Ramage, and Roselander]{fl_at_scale}
Kallista Bonawitz, Hubert Eichner, Wolfgang Grieskamp, Dzmitry Huba, Alex Ingerman, Vladimir Ivanov, Chlo\'{e} Kiddon, Jakub Kone\v{c}n\'{y}, Stefano Mazzocchi, Brendan McMahan, Timon Van~Overveldt, David Petrou, Daniel Ramage, and Jason Roselander.
\newblock Towards federated learning at scale: System design.
\newblock In A.~Talwalkar, V.~Smith, and M.~Zaharia, editors, \emph{Proceedings of Machine Learning and Systems}, volume~1, pages 374--388, 2019.
\newblock URL \url{https://proceedings.mlsys.org/paper/2019/file/bd686fd640be98efaae0091fa301e613-Paper.pdf}.

\bibitem[Bradbury et~al.(2018)Bradbury, Frostig, Hawkins, Johnson, Leary, Maclaurin, Necula, Paszke, VanderPlas, Wanderman-Milne, et~al.]{jax}
James Bradbury, Roy Frostig, Peter Hawkins, Matthew~James Johnson, Chris Leary, Dougal Maclaurin, George Necula, Adam Paszke, Jake VanderPlas, Skye Wanderman-Milne, et~al.
\newblock {JAX}: composable transformations of {Python}+ {NumPy} programs.
\newblock \emph{Version 0.2}, 5:\penalty0 14--24, 2018.

\bibitem[Charles and Kone\v{c}n\'y(2021)]{charles_and_konecny}
Zachary Charles and Jakub Kone\v{c}n\'y.
\newblock Convergence and accuracy trade-offs in federated learning and meta-learning.
\newblock In Arindam Banerjee and Kenji Fukumizu, editors, \emph{Proceedings of The 24th International Conference on Artificial Intelligence and Statistics}, volume 130 of \emph{Proceedings of Machine Learning Research}, pages 2575--2583. PMLR, 13--15 Apr 2021.
\newblock URL \url{https://proceedings.mlr.press/v130/charles21a.html}.

\bibitem[Charles and Rush(2022)]{charles_and_rush}
Zachary Charles and Keith Rush.
\newblock Iterated vector fields and conservatism, with applications to federated learning.
\newblock In Sanjoy Dasgupta and Nika Haghtalab, editors, \emph{Proceedings of The 33rd International Conference on Algorithmic Learning Theory}, volume 167 of \emph{Proceedings of Machine Learning Research}, pages 130--147. PMLR, 29 Mar--01 Apr 2022.
\newblock URL \url{https://proceedings.mlr.press/v167/charles22a.html}.

\bibitem[Charles et~al.(2021)Charles, Garrett, Huo, Shmulyian, and Smith]{large_cohort}
Zachary Charles, Zachary Garrett, Zhouyuan Huo, Sergei Shmulyian, and Virginia Smith.
\newblock On large-cohort training for federated learning.
\newblock In M.~Ranzato, A.~Beygelzimer, Y.~Dauphin, P.S. Liang, and J.~Wortman Vaughan, editors, \emph{Advances in Neural Information Processing Systems}, volume~34, pages 20461--20475. Curran Associates, Inc., 2021.
\newblock URL \url{https://proceedings.neurips.cc/paper/2021/file/ab9ebd57177b5106ad7879f0896685d4-Paper.pdf}.

\bibitem[Charles et~al.(2022)Charles, Bonawitz, Chiknavaryan, McMahan, et~al.]{charles2022federated}
Zachary Charles, Kallista Bonawitz, Stanislav Chiknavaryan, Brendan McMahan, et~al.
\newblock Federated select: A primitive for communication-and memory-efficient federated learning.
\newblock \emph{arXiv preprint arXiv:2208.09432}, 2022.

\bibitem[Chen et~al.(2019)Chen, Mathews, Ouyang, and Beaufays]{oov_fl}
Mingqing Chen, Rajiv Mathews, Tom Ouyang, and Fran{\c{c}}oise Beaufays.
\newblock Federated learning of out-of-vocabulary words.
\newblock \emph{CoRR}, abs/1903.10635, 2019.
\newblock URL \url{http://arxiv.org/abs/1903.10635}.

\bibitem[Christianson(1994)]{christianson1994reverse}
Bruce Christianson.
\newblock Reverse accumulation and attractive fixed points.
\newblock \emph{Optimization Methods and Software}, 3\penalty0 (4):\penalty0 311--326, 1994.

\bibitem[Cohen et~al.(2017)Cohen, Afshar, Tapson, and Van~Schaik]{cohen2017emnist}
Gregory Cohen, Saeed Afshar, Jonathan Tapson, and Andre Van~Schaik.
\newblock {EMNIST}: Extending {MNIST} to handwritten letters.
\newblock In \emph{2017 International Joint Conference on Neural Networks (IJCNN)}, pages 2921--2926. IEEE, 2017.

\bibitem[Elsken et~al.(2019)Elsken, Metzen, and Hutter]{elsken2019neural}
Thomas Elsken, Jan~Hendrik Metzen, and Frank Hutter.
\newblock Neural architecture search: {A} survey.
\newblock \emph{J. Mach. Learn. Res.}, 20:\penalty0 55:1--55:21, 2019.
\newblock URL \url{http://jmlr.org/papers/v20/18-598.html}.

\bibitem[Fallah et~al.(2020)Fallah, Mokhtari, and Ozdaglar]{perfedavg}
Alireza Fallah, Aryan Mokhtari, and Asuman Ozdaglar.
\newblock Personalized federated learning with theoretical guarantees: A model-agnostic meta-learning approach.
\newblock \emph{Advances in Neural Information Processing Systems}, 33:\penalty0 3557--3568, 2020.

\bibitem[Finn et~al.(2017)Finn, Abbeel, and Levine]{maml}
Chelsea Finn, Pieter Abbeel, and Sergey Levine.
\newblock Model-agnostic meta-learning for fast adaptation of deep networks.
\newblock In Doina Precup and Yee~Whye Teh, editors, \emph{Proceedings of the 34th International Conference on Machine Learning}, volume~70 of \emph{Proceedings of Machine Learning Research}, pages 1126--1135. PMLR, 06--11 Aug 2017.
\newblock URL \url{https://proceedings.mlr.press/v70/finn17a.html}.

\bibitem[Griewank and Walther(2008)]{principles_of_autodiff}
Andreas Griewank and Andrea Walther.
\newblock \emph{Evaluating derivatives - principles and techniques of algorithmic differentiation, Second Edition}.
\newblock {SIAM}, 2008.
\newblock ISBN 978-0-89871-659-7.
\newblock \doi{10.1137/1.9780898717761}.
\newblock URL \url{https://doi.org/10.1137/1.9780898717761}.

\bibitem[Habiba and Pearlmutter(2021)]{habiba2021neural}
Mansura Habiba and Barak~A Pearlmutter.
\newblock Neural network based on automatic differentiation transformation of numeric iterate-to-fixedpoint.
\newblock In \emph{2021 International Conference on Electrical, Computer and Energy Technologies (ICECET)}, pages 1--6. IEEE, 2021.

\bibitem[Hard et~al.(2018)Hard, Rao, Mathews, Beaufays, Augenstein, Eichner, Kiddon, and Ramage]{NWP18}
Andrew Hard, Kanishka Rao, Rajiv Mathews, Fran{\c{c}}oise Beaufays, Sean Augenstein, Hubert Eichner, Chlo{\'{e}} Kiddon, and Daniel Ramage.
\newblock Federated learning for mobile keyboard prediction.
\newblock \emph{CoRR}, abs/1811.03604, 2018.
\newblock URL \url{http://arxiv.org/abs/1811.03604}.

\bibitem[Hard et~al.(2020)Hard, Partridge, Nguyen, Subrahmanya, Shah, Zhu, L{\'{o}}pez{-}Moreno, and Mathews]{keyword_speech}
Andrew Hard, Kurt Partridge, Cameron Nguyen, Niranjan Subrahmanya, Aishanee Shah, Pai Zhu, Ignacio L{\'{o}}pez{-}Moreno, and Rajiv Mathews.
\newblock Training keyword spotting models on non-{IID} data with federated learning.
\newblock In Helen Meng, Bo~Xu, and Thomas~Fang Zheng, editors, \emph{Interspeech 2020, 21st Annual Conference of the International Speech Communication Association, Virtual Event, Shanghai, China, 25-29 October 2020}, pages 4343--4347. {ISCA}, 2020.
\newblock \doi{10.21437/INTERSPEECH.2020-3023}.
\newblock URL \url{https://doi.org/10.21437/Interspeech.2020-3023}.

\bibitem[Hou et~al.(2021)Hou, Thekumparampil, Fanti, and Oh]{fedchain}
Charlie Hou, Kiran~Koshy Thekumparampil, Giulia Fanti, and Sewoong Oh.
\newblock Fed{C}hain: Chained algorithms for near-optimal communication cost in federated learning.
\newblock In \emph{International Conference on Learning Representations}, 2021.

\bibitem[Hsieh et~al.(2020)Hsieh, Phanishayee, Mutlu, and Gibbons]{groupnorm_for_fl}
Kevin Hsieh, Amar Phanishayee, Onur Mutlu, and Phillip Gibbons.
\newblock The non-{IID} data quagmire of decentralized machine learning.
\newblock In \emph{International Conference on Machine Learning}, pages 4387--4398. PMLR, 2020.

\bibitem[Hsu et~al.(2019)Hsu, Qi, and Brown]{fedavgm}
Tzu{-}Ming~Harry Hsu, Hang Qi, and Matthew Brown.
\newblock Measuring the effects of non-identical data distribution for federated visual classification.
\newblock \emph{CoRR}, abs/1909.06335, 2019.
\newblock URL \url{http://arxiv.org/abs/1909.06335}.

\bibitem[Huba et~al.(2022)Huba, Nguyen, Malik, Zhu, Rabbat, Yousefpour, Wu, Zhan, Ustinov, Srinivas, Wang, Shoumikhin, Min, and Malek]{papaya}
Dzmitry Huba, John Nguyen, Kshitiz Malik, Ruiyu Zhu, Mike Rabbat, Ashkan Yousefpour, Carole{-}Jean Wu, Hongyuan Zhan, Pavel Ustinov, Harish Srinivas, Kaikai Wang, Anthony Shoumikhin, Jesik Min, and Mani Malek.
\newblock {PAPAYA:} practical, private, and scalable federated learning.
\newblock In Diana Marculescu, Yuejie Chi, and Carole{-}Jean Wu, editors, \emph{Proceedings of Machine Learning and Systems 2022, MLSys 2022, Santa Clara, CA, USA, August 29 - September 1, 2022}. mlsys.org, 2022.
\newblock URL \url{https://proceedings.mlsys.org/paper/2022/hash/f340f1b1f65b6df5b5e3f94d95b11daf-Abstract.html}.

\bibitem[Ingerman and Ostrowski(2019)]{ingerman_ostrowski_2019}
Alex Ingerman and Krzysztof Ostrowski.
\newblock Introducing {T}ensor{F}low {F}ederated, Mar 2019.
\newblock URL \url{https://blog.tensorflow.org/2019/03/introducing-tensorflow-federated.html}.

\bibitem[Kairouz et~al.(2021)Kairouz, McMahan, Avent, Bellet, Bennis, Bhagoji, Bonawitz, Charles, Cormode, Cummings, D'Oliveira, Eichner, Rouayheb, Evans, Gardner, Garrett, Gasc{\'{o}}n, Ghazi, Gibbons, Gruteser, Harchaoui, He, He, Huo, Hutchinson, Hsu, Jaggi, Javidi, Joshi, Khodak, Kone{\v{c}}n{\'y}, Korolova, Koushanfar, Koyejo, Lepoint, Liu, Mittal, Mohri, Nock, {\"{O}}zg{\"{u}}r, Pagh, Qi, Ramage, Raskar, Raykova, Song, Song, Stich, Sun, Suresh, Tram{\`{e}}r, Vepakomma, Wang, Xiong, Xu, Yang, Yu, Yu, and Zhao]{aop}
Peter Kairouz, H.~Brendan McMahan, Brendan Avent, Aur{\'{e}}lien Bellet, Mehdi Bennis, Arjun~Nitin Bhagoji, Kallista~A. Bonawitz, Zachary Charles, Graham Cormode, Rachel Cummings, Rafael G.~L. D'Oliveira, Hubert Eichner, Salim~El Rouayheb, David Evans, Josh Gardner, Zachary Garrett, Adri{\`{a}} Gasc{\'{o}}n, Badih Ghazi, Phillip~B. Gibbons, Marco Gruteser, Za{\"{\i}}d Harchaoui, Chaoyang He, Lie He, Zhouyuan Huo, Ben Hutchinson, Justin Hsu, Martin Jaggi, Tara Javidi, Gauri Joshi, Mikhail Khodak, Jakub Kone{\v{c}}n{\'y}, Aleksandra Korolova, Farinaz Koushanfar, Sanmi Koyejo, Tancr{\`{e}}de Lepoint, Yang Liu, Prateek Mittal, Mehryar Mohri, Richard Nock, Ayfer {\"{O}}zg{\"{u}}r, Rasmus Pagh, Hang Qi, Daniel Ramage, Ramesh Raskar, Mariana Raykova, Dawn Song, Weikang Song, Sebastian~U. Stich, Ziteng Sun, Ananda~Theertha Suresh, Florian Tram{\`{e}}r, Praneeth Vepakomma, Jianyu Wang, Li~Xiong, Zheng Xu, Qiang Yang, Felix~X. Yu, Han Yu, and Sen Zhao.
\newblock Advances and open problems in federated learning.
\newblock \emph{Found. Trends Mach. Learn.}, 14\penalty0 (1-2):\penalty0 1--210, 2021.
\newblock \doi{10.1561/2200000083}.
\newblock URL \url{https://doi.org/10.1561/2200000083}.

\bibitem[Kan(2022)]{fed_hypergrad_descent}
Andrew~K. Kan.
\newblock Federated hypergradient descent.
\newblock \emph{CoRR}, abs/2211.02106, 2022.
\newblock \doi{10.48550/ARXIV.2211.02106}.
\newblock URL \url{https://doi.org/10.48550/arXiv.2211.02106}.

\bibitem[Karimireddy et~al.(2020)Karimireddy, Kale, Mohri, Reddi, Stich, and Suresh]{scaffold}
Sai~Praneeth Karimireddy, Satyen Kale, Mehryar Mohri, Sashank Reddi, Sebastian Stich, and Ananda~Theertha Suresh.
\newblock Scaffold: Stochastic controlled averaging for federated learning.
\newblock In \emph{International Conference on Machine Learning}, pages 5132--5143. PMLR, 2020.

\bibitem[Khodak et~al.(2021)Khodak, Tu, Li, Li, Balcan, Smith, and Talwalkar]{fed_hparam_tuning}
Mikhail Khodak, Renbo Tu, Tian Li, Liam Li, Maria-Florina~F Balcan, Virginia Smith, and Ameet Talwalkar.
\newblock Federated hyperparameter tuning: Challenges, baselines, and connections to weight-sharing.
\newblock In M.~Ranzato, A.~Beygelzimer, Y.~Dauphin, P.S. Liang, and J.~Wortman Vaughan, editors, \emph{Advances in Neural Information Processing Systems}, volume~34, pages 19184--19197. Curran Associates, Inc., 2021.
\newblock URL \url{https://proceedings.neurips.cc/paper/2021/file/a0205b87490c847182672e8d371e9948-Paper.pdf}.

\bibitem[Kingma and Ba(2015)]{adam}
Diederik~P. Kingma and Jimmy Ba.
\newblock Adam: {A} method for stochastic optimization.
\newblock In Yoshua Bengio and Yann LeCun, editors, \emph{3rd International Conference on Learning Representations, {ICLR} 2015, San Diego, CA, USA, May 7-9, 2015, Conference Track Proceedings}, 2015.
\newblock URL \url{http://arxiv.org/abs/1412.6980}.

\bibitem[Krizhevsky(2009)]{krizhevsky2009learning}
Alex Krizhevsky.
\newblock Learning multiple layers of features from tiny images.
\newblock Technical report, 2009.

\bibitem[Li and Malik(2016)]{li17a}
Ke~Li and Jitendra Malik.
\newblock Learning to optimize.
\newblock \emph{arXiv preprint arXiv:1606.01885}, 2016.

\bibitem[Li and Malik(2017)]{li17b}
Ke~Li and Jitendra Malik.
\newblock Learning to optimize neural nets.
\newblock \emph{arXiv preprint arXiv:1703.00441}, 2017.

\bibitem[Li et~al.(2020)Li, Sahu, Zaheer, Sanjabi, Talwalkar, and Smith]{fedprox}
Tian Li, Anit~Kumar Sahu, Manzil Zaheer, Maziar Sanjabi, Ameet Talwalkar, and Virginia Smith.
\newblock Federated optimization in heterogeneous networks.
\newblock \emph{Proceedings of Machine Learning and Systems}, 2:\penalty0 429--450, 2020.

\bibitem[Lv et~al.(2017)Lv, Jiang, and Li]{lv17}
Kaifeng Lv, Shunhua Jiang, and Jian Li.
\newblock Learning gradient descent: Better generalization and longer horizons.
\newblock In \emph{International Conference on Machine Learning}, pages 2247--2255. PMLR, 2017.

\bibitem[Malinovskiy et~al.(2020)Malinovskiy, Kovalev, Gasanov, Condat, and Richtarik]{local_fixed_points_kaust}
Grigory Malinovskiy, Dmitry Kovalev, Elnur Gasanov, Laurent Condat, and Peter Richtarik.
\newblock From local {SGD} to local fixed-point methods for federated learning.
\newblock In Hal~Daumé III and Aarti Singh, editors, \emph{Proceedings of the 37th International Conference on Machine Learning}, volume 119 of \emph{Proceedings of Machine Learning Research}, pages 6692--6701. PMLR, 13--18 Jul 2020.
\newblock URL \url{https://proceedings.mlr.press/v119/malinovskiy20a.html}.

\bibitem[Manzyuk et~al.(2019)Manzyuk, Pearlmutter, Radul, Rush, and Siskind]{manzyuk2019perturbation}
Oleksandr Manzyuk, Barak~A Pearlmutter, Alexey~Andreyevich Radul, David~R Rush, and Jeffrey~Mark Siskind.
\newblock Perturbation confusion in forward automatic differentiation of higher-order functions.
\newblock \emph{Journal of Functional Programming}, 29:\penalty0 e12, 2019.

\bibitem[McMahan et~al.(2017)McMahan, Moore, Ramage, Hampson, and y~Arcas]{mcmahan2017communication}
Brendan McMahan, Eider Moore, Daniel Ramage, Seth Hampson, and Blaise~Aguera y~Arcas.
\newblock Communication-efficient learning of deep networks from decentralized data.
\newblock In \emph{Artificial intelligence and statistics}, pages 1273--1282. PMLR, 2017.

\bibitem[Metz et~al.(2019{\natexlab{a}})Metz, Maheswaranathan, Cheung, and Sohl{-}Dickstein]{meta_learning_unsupervised}
Luke Metz, Niru Maheswaranathan, Brian Cheung, and Jascha Sohl{-}Dickstein.
\newblock Meta-learning update rules for unsupervised representation learning.
\newblock In \emph{7th International Conference on Learning Representations, {ICLR} 2019, New Orleans, LA, USA, May 6-9, 2019}. OpenReview.net, 2019{\natexlab{a}}.
\newblock URL \url{https://openreview.net/forum?id=HkNDsiC9KQ}.

\bibitem[Metz et~al.(2019{\natexlab{b}})Metz, Maheswaranathan, Nixon, Freeman, and Sohl-Dickstein]{metz19a}
Luke Metz, Niru Maheswaranathan, Jeremy Nixon, Daniel Freeman, and Jascha Sohl-Dickstein.
\newblock Understanding and correcting pathologies in the training of learned optimizers.
\newblock In Kamalika Chaudhuri and Ruslan Salakhutdinov, editors, \emph{Proceedings of the 36th International Conference on Machine Learning}, volume~97 of \emph{Proceedings of Machine Learning Research}, pages 4556--4565. PMLR, 09--15 Jun 2019{\natexlab{b}}.
\newblock URL \url{https://proceedings.mlr.press/v97/metz19a.html}.

\bibitem[Metz et~al.(2022)Metz, Harrison, Freeman, Merchant, Beyer, Bradbury, Agrawal, Poole, Mordatch, Roberts, and Sohl{-}Dickstein]{velo_metz}
Luke Metz, James Harrison, C.~Daniel Freeman, Amil Merchant, Lucas Beyer, James Bradbury, Naman Agrawal, Ben Poole, Igor Mordatch, Adam Roberts, and Jascha Sohl{-}Dickstein.
\newblock Velo: Training versatile learned optimizers by scaling up.
\newblock \emph{CoRR}, abs/2211.09760, 2022.
\newblock \doi{10.48550/ARXIV.2211.09760}.
\newblock URL \url{https://doi.org/10.48550/arXiv.2211.09760}.

\bibitem[Mishchenko et~al.(2022)Mishchenko, Malinovsky, Stich, and Richt{\'{a}}rik]{proxskip}
Konstantin Mishchenko, Grigory Malinovsky, Sebastian~U. Stich, and Peter Richt{\'{a}}rik.
\newblock Prox{S}kip: Yes! local gradient steps provably lead to communication acceleration! finally!
\newblock In Kamalika Chaudhuri, Stefanie Jegelka, Le~Song, Csaba Szepesv{\'{a}}ri, Gang Niu, and Sivan Sabato, editors, \emph{International Conference on Machine Learning, {ICML} 2022, 17-23 July 2022, Baltimore, Maryland, {USA}}, volume 162 of \emph{Proceedings of Machine Learning Research}, pages 15750--15769. {PMLR}, 2022.
\newblock URL \url{https://proceedings.mlr.press/v162/mishchenko22b.html}.

\bibitem[Mitra et~al.(2021)Mitra, Jaafar, Pappas, and Hassani]{fedlin}
Aritra Mitra, Rayana~H. Jaafar, George~J. Pappas, and Hamed Hassani.
\newblock Linear convergence in federated learning: Tackling client heterogeneity and sparse gradients.
\newblock In Marc'Aurelio Ranzato, Alina Beygelzimer, Yann~N. Dauphin, Percy Liang, and Jennifer~Wortman Vaughan, editors, \emph{Advances in Neural Information Processing Systems 34: Annual Conference on Neural Information Processing Systems 2021, NeurIPS 2021, December 6-14, 2021, virtual}, pages 14606--14619, 2021.
\newblock URL \url{https://proceedings.neurips.cc/paper/2021/hash/7a6bda9ad6ffdac035c752743b7e9d0e-Abstract.html}.

\bibitem[Naumann(2008)]{oja}
Uwe Naumann.
\newblock Optimal jacobian accumulation is np-complete.
\newblock \emph{Math. Program.}, 112\penalty0 (2):\penalty0 427–441, apr 2008.
\newblock ISSN 0025-5610.

\bibitem[Oktay et~al.(2020)Oktay, Ball{\'{e}}, Singh, and Shrivastava]{Oktay2020Scalable}
Deniz Oktay, Johannes Ball{\'{e}}, Saurabh Singh, and Abhinav Shrivastava.
\newblock Scalable model compression by entropy penalized reparameterization.
\newblock In \emph{8th International Conference on Learning Representations, {ICLR} 2020, Addis Ababa, Ethiopia, April 26-30, 2020}. OpenReview.net, 2020.
\newblock URL \url{https://openreview.net/forum?id=HkgxW0EYDS}.

\bibitem[Paszke et~al.(2019)Paszke, Gross, Massa, Lerer, Bradbury, Chanan, Killeen, Lin, Gimelshein, Antiga, et~al.]{pytorch}
Adam Paszke, Sam Gross, Francisco Massa, Adam Lerer, James Bradbury, Gregory Chanan, Trevor Killeen, Zeming Lin, Natalia Gimelshein, Luca Antiga, et~al.
\newblock {PyT}orch: An imperative style, high-performance deep learning library.
\newblock \emph{Advances in neural information processing systems}, 32, 2019.

\bibitem[Pathak and Wainwright(2020)]{fedsplit}
Reese Pathak and Martin~J. Wainwright.
\newblock Fedsplit: an algorithmic framework for fast federated optimization.
\newblock In Hugo Larochelle, Marc'Aurelio Ranzato, Raia Hadsell, Maria{-}Florina Balcan, and Hsuan{-}Tien Lin, editors, \emph{Advances in Neural Information Processing Systems 33: Annual Conference on Neural Information Processing Systems 2020, NeurIPS 2020, December 6-12, 2020, virtual}, 2020.
\newblock URL \url{https://proceedings.neurips.cc/paper/2020/hash/4ebd440d99504722d80de606ea8507da-Abstract.html}.

\bibitem[Paulik et~al.(2021)Paulik, Seigel, Mason, Telaar, Kluivers, van Dalen, Lau, Carlson, Granqvist, Vandevelde, Agarwal, Freudiger, Byde, Bhowmick, Kapoor, Beaumont, Cahill, Hughes, Javidbakht, Dong, Rishi, and Hung]{apple_fl}
Matthias Paulik, Matt Seigel, Henry Mason, Dominic Telaar, Joris Kluivers, Rogier~C. van Dalen, Chi~Wai Lau, Luke Carlson, Filip Granqvist, Chris Vandevelde, Sudeep Agarwal, Julien Freudiger, Andrew Byde, Abhishek Bhowmick, Gaurav Kapoor, Si~Beaumont, {\'{A}}ine Cahill, Dominic Hughes, Omid Javidbakht, Fei Dong, Rehan Rishi, and Stanley Hung.
\newblock Federated evaluation and tuning for on-device personalization: System design {\&} applications.
\newblock \emph{CoRR}, abs/2102.08503, 2021.
\newblock URL \url{https://arxiv.org/abs/2102.08503}.

\bibitem[Pham et~al.(2018)Pham, Guan, Zoph, Le, and Dean]{pham2018efficient}
Hieu Pham, Melody~Y. Guan, Barret Zoph, Quoc~V. Le, and Jeff Dean.
\newblock Efficient neural architecture search via parameter sharing.
\newblock In Jennifer~G. Dy and Andreas Krause, editors, \emph{Proceedings of the 35th International Conference on Machine Learning, {ICML} 2018, Stockholmsm{\"{a}}ssan, Stockholm, Sweden, July 10-15, 2018}, volume~80 of \emph{Proceedings of Machine Learning Research}, pages 4092--4101. {PMLR}, 2018.
\newblock URL \url{http://proceedings.mlr.press/v80/pham18a.html}.

\bibitem[Pillutla et~al.(2022)Pillutla, Kakade, and Harchaoui]{robust_agg}
Krishna Pillutla, Sham~M. Kakade, and Za{\"{\i}}d Harchaoui.
\newblock Robust aggregation for federated learning.
\newblock \emph{{IEEE} Trans. Signal Process.}, 70:\penalty0 1142--1154, 2022.
\newblock \doi{10.1109/TSP.2022.3153135}.
\newblock URL \url{https://doi.org/10.1109/TSP.2022.3153135}.

\bibitem[Ramaswamy et~al.(2019)Ramaswamy, Mathews, Rao, and Beaufays]{emoji_fl}
Swaroop Ramaswamy, Rajiv Mathews, Kanishka Rao, and Fran{\c{c}}oise Beaufays.
\newblock Federated learning for emoji prediction in a mobile keyboard.
\newblock \emph{CoRR}, abs/1906.04329, 2019.
\newblock URL \url{http://arxiv.org/abs/1906.04329}.

\bibitem[Reddi et~al.(2021)Reddi, Charles, Zaheer, Garrett, Rush, Kone{\v{c}}n{\'y}, Kumar, and McMahan]{afo}
Sashank~J. Reddi, Zachary Charles, Manzil Zaheer, Zachary Garrett, Keith Rush, Jakub Kone{\v{c}}n{\'y}, Sanjiv Kumar, and Hugh~Brendan McMahan.
\newblock Adaptive federated optimization.
\newblock In \emph{International Conference on Learning Representations}, 2021.
\newblock URL \url{https://openreview.net/forum?id=LkFG3lB13U5}.

\bibitem[Rush et~al.(2024)Rush, Charles, and Garrett]{rush2024fax}
Keith Rush, Zachary Charles, and Zachary Garrett.
\newblock Fax: Scalable and differentiable federated primitives in jax.
\newblock \emph{arXiv preprint arXiv:2403.07128}, 2024.

\bibitem[Rusu et~al.(2019)Rusu, Rao, Sygnowski, Vinyals, Pascanu, Osindero, and Hadsell]{meta_embed}
Andrei~A. Rusu, Dushyant Rao, Jakub Sygnowski, Oriol Vinyals, Razvan Pascanu, Simon Osindero, and Raia Hadsell.
\newblock Meta-learning with latent embedding optimization.
\newblock 2019.
\newblock URL \url{https://openreview.net/forum?id=BJgklhAcK7}.

\bibitem[Sandler et~al.(2021)Sandler, Vladymyrov, Zhmoginov, Miller, Madams, Jackson, and y~Arcas]{blur}
Mark Sandler, Max Vladymyrov, Andrey Zhmoginov, Nolan Miller, Tom Madams, Andrew Jackson, and Blaise~Ag{\"{u}}era y~Arcas.
\newblock Meta-learning bidirectional update rules.
\newblock In Marina Meila and Tong Zhang, editors, \emph{Proceedings of the 38th International Conference on Machine Learning, {ICML} 2021, 18-24 July 2021, Virtual Event}, volume 139 of \emph{Proceedings of Machine Learning Research}, pages 9288--9300. {PMLR}, 2021.
\newblock URL \url{http://proceedings.mlr.press/v139/sandler21a.html}.

\bibitem[Schmidhuber(1994)]{Schmidhuber1994OnLH}
J~Schmidhuber.
\newblock On learning how to learn learning strategies (technical report {FKI}-198-94).
\newblock \emph{Fakultat fur Informatik}, 1994.

\bibitem[Siskind and Pearlmutter(2005)]{siskind2005perturbation}
Jeffrey~Mark Siskind and Barak~A Pearlmutter.
\newblock Perturbation confusion and referential transparency: Correct functional implementation of forward-mode ad.
\newblock 2005.

\bibitem[Tarzanagh et~al.(2022)Tarzanagh, Li, Thrampoulidis, and Oymak]{fednest}
Davoud~Ataee Tarzanagh, Mingchen Li, Christos Thrampoulidis, and Samet Oymak.
\newblock Fed{N}est: Federated bilevel, minimax, and compositional optimization.
\newblock In Kamalika Chaudhuri, Stefanie Jegelka, Le~Song, Csaba Szepesv{\'{a}}ri, Gang Niu, and Sivan Sabato, editors, \emph{International Conference on Machine Learning, {ICML} 2022, 17-23 July 2022, Baltimore, Maryland, {USA}}, volume 162 of \emph{Proceedings of Machine Learning Research}, pages 21146--21179. {PMLR}, 2022.
\newblock URL \url{https://proceedings.mlr.press/v162/tarzanagh22a.html}.

\bibitem[Vicol et~al.(2021)Vicol, Metz, and Sohl-Dickstein]{vicol21a}
Paul Vicol, Luke Metz, and Jascha Sohl-Dickstein.
\newblock Unbiased gradient estimation in unrolled computation graphs with persistent evolution strategies.
\newblock In Marina Meila and Tong Zhang, editors, \emph{Proceedings of the 38th International Conference on Machine Learning}, volume 139 of \emph{Proceedings of Machine Learning Research}, pages 10553--10563. PMLR, 18--24 Jul 2021.
\newblock URL \url{https://proceedings.mlr.press/v139/vicol21a.html}.

\bibitem[Wang et~al.(2020)Wang, Liu, Liang, Joshi, and Poor]{fednova}
Jianyu Wang, Qinghua Liu, Hao Liang, Gauri Joshi, and H~Vincent Poor.
\newblock Tackling the objective inconsistency problem in heterogeneous federated optimization.
\newblock \emph{Advances in neural information processing systems}, 33:\penalty0 7611--7623, 2020.

\bibitem[Wang et~al.(2023)Wang, Wang, and Li]{wang2023fedhyper}
Ziyao Wang, Jianyu Wang, and Ang Li.
\newblock Fed{H}yper: A universal and robust learning rate scheduler for federated learning with hypergradient descent.
\newblock In \emph{The Twelfth International Conference on Learning Representations}, 2023.

\bibitem[Wichrowska et~al.(2017)Wichrowska, Maheswaranathan, Hoffman, Colmenarejo, Denil, de~Freitas, and Sohl-Dickstein]{wichrowska17a}
Olga Wichrowska, Niru Maheswaranathan, Matthew~W. Hoffman, Sergio~G{\'o}mez Colmenarejo, Misha Denil, Nando de~Freitas, and Jascha Sohl-Dickstein.
\newblock Learned optimizers that scale and generalize.
\newblock In Doina Precup and Yee~Whye Teh, editors, \emph{Proceedings of the 34th International Conference on Machine Learning}, volume~70 of \emph{Proceedings of Machine Learning Research}, pages 3751--3760. PMLR, 06--11 Aug 2017.
\newblock URL \url{https://proceedings.mlr.press/v70/wichrowska17a.html}.

\bibitem[Yang et~al.(2018)Yang, Andrew, Eichner, Sun, Li, Kong, Ramage, and Beaufays]{gboard_18}
Timothy Yang, Galen Andrew, Hubert Eichner, Haicheng Sun, Wei Li, Nicholas Kong, Daniel Ramage, and Fran{\c{c}}oise Beaufays.
\newblock Applied federated learning: Improving google keyboard query suggestions.
\newblock \emph{CoRR}, abs/1812.02903, 2018.
\newblock URL \url{http://arxiv.org/abs/1812.02903}.

\bibitem[Yin et~al.(2018)Yin, Chen, Kannan, and Bartlett]{yin2018byzantine}
Dong Yin, Yudong Chen, Ramchandran Kannan, and Peter Bartlett.
\newblock Byzantine-robust distributed learning: Towards optimal statistical rates.
\newblock In \emph{International Conference on Machine Learning}, pages 5650--5659. PMLR, 2018.

\bibitem[Zoph and Le(2017)]{zoph2016neural}
Barret Zoph and Quoc~V. Le.
\newblock Neural architecture search with reinforcement learning.
\newblock In \emph{5th International Conference on Learning Representations, {ICLR} 2017, Toulon, France, April 24-26, 2017, Conference Track Proceedings}. OpenReview.net, 2017.
\newblock URL \url{https://openreview.net/forum?id=r1Ue8Hcxg}.

\end{thebibliography}

\end{document}